%% file: main.tex
\documentclass[10pt,twocolumn,letterpaper]{article}

\usepackage{3dv}
\usepackage{times}
\usepackage{epsfig}
\usepackage{graphicx}
\usepackage{amsmath}
\usepackage{amssymb}
\usepackage{enumitem} % for itemized bullets
\usepackage{euscript} % for curly N
\usepackage{bbm} % for indicator function 1
\usepackage{multirow} % for multirow in a table
\usepackage{booktabs} % for using top, mid, and bottom rule
\usepackage[pagebackref=true,breaklinks=true,letterpaper=true,colorlinks,bookmarks=false]{hyperref}

\makeatletter
\newcommand{\printfnsymbol}[1]{%
  \textsuperscript{\@fnsymbol{#1}}%
}
\makeatother

\makeatletter
\renewcommand\@makefntext[1]{\leftskip=0em\hskip-0em\@makefnmark#1}
\makeatother

\threedvfinalcopy % *** Uncomment this line for the final submission

\def\threedvPaperID{52} % *** Enter the 3DV Paper ID here

\ifthreedvfinal\fi
\input{macros.tex}

\newcommand{\Xcal}{\mathcal{X}}
\newcommand{\Ycal}{\mathcal{Y}}
\newcommand{\Pcal}{\mathcal{P}}
\newcommand{\Qcal}{\mathcal{Q}}
\newcommand{\Lcal}{\mathcal{L}}
\newcommand{\Neu}{\EuScript{N}}

\DeclareMathOperator*{\argmax}{\arg\!\max}

\begin{document}

%%%%%%%%% TITLE
\title{DPC: Unsupervised Deep Point Correspondence via Cross and Self Construction}

\input{authors_one_line.tex}

\maketitle
\thispagestyle{empty} % *** Uncomment this line for the final submission

%-------------------------------------------------------------------------
%%%%%%%%% ABSTRACT
\input{00_abstract.tex}
\input{01_introduction.tex}

%-------------------------------------------------------------------------
%%%%%%%%% RELATED WORK
\input{02_related_work.tex}
\input{03_method.tex}

%-------------------------------------------------------------------------
%%%%%%%%% RESULTS
\input{04_results.tex}

%-------------------------------------------------------------------------
%%%%%%%%% CONCLUSIONS
\input{05_conclusions.tex}

%-------------------------------------------------------------------------
%%%%%%%%% REFERENCES
{\small
\bibliographystyle{ieee}
\bibliography{references.bib}
}

%-------------------------------------------------------------------------
%%%%%%%%% SUPPLEMENTARY MATERIAL
\input{supplementary/supplementary.tex}

\end{document}

%% file: macros.tex
\long\def\ignorethis#1{}

\usepackage{color}
\definecolor{gray}{rgb}{0.6,0.6,0.6}
\definecolor{red}{rgb}{1,0,0}
\definecolor{green}{rgb}{0,1,0}
\definecolor{blue}{rgb}{0,0,1}
\definecolor{dark-green}{rgb}{0,0.4,0}
\definecolor{orange}{rgb}{1,0.55,0}
\definecolor{white}{rgb}{1,1,1}
\definecolor{black}{rgb}{0,0,0}
\definecolor{dark-brown}{rgb}{0.2,0.1,0}
\definecolor{light-blue}{rgb}{0.4,0.6,0.99}
\definecolor{dark-red}{rgb}{0.6,0,0}
\definecolor{light-red}{rgb}{1,0.2,0.6}
\definecolor{pink}{rgb}{1,0.2,0.6}
\definecolor{dark-pink}{rgb}{0.6,0,0.3}

\newcommand{\graytxt}[1]{{\color{gray}#1}\normalfont}
\newcommand{\whitetext}[1]{{\color{white}#1}\normalfont}

\newbox\jsavebox

%% file: authors_one_line.tex
\author{
% Name
Itai Lang\textsuperscript{*} \qquad \whitetext{aaaaaa}Dvir Ginzburg\textsuperscript{*} \qquad \whitetext{aaaaa}Shai Avidan \qquad \whitetext{aaaaaaa}Dan Raviv \\
% Institution
Tel Aviv University \qquad Tel Aviv University \qquad Tel Aviv University \qquad Tel Aviv University \\
% Email
{\tt\small \hspace{3mm} \{itailang, dvirginzburg\}@mail.tau.ac.il \hspace{4.5mm} avidan@eng.tau.ac.il \hspace{0.5mm} darav@tauex.tau.ac.il}
}

%% file: 00_abstract.tex
\begin{abstract}
We present a new method for real-time non-rigid dense correspondence between point clouds based on structured shape construction. Our method, termed Deep Point Correspondence (DPC), requires a fraction of the training data compared to previous techniques and presents better generalization capabilities.
Until now, two main approaches have been suggested for the dense correspondence problem. The first is a spectral-based approach that obtains great results on synthetic datasets but requires mesh connectivity of the shapes and long inference processing time while being unstable in real-world scenarios. The second is a spatial approach that uses an encoder-decoder framework to regress an ordered point cloud for the matching alignment from an irregular input. Unfortunately, the decoder brings considerable disadvantages, as it requires a large amount of training data and struggles to generalize well in cross-dataset evaluations.
DPC's novelty lies in its \emph{lack} of a decoder component. Instead, we use latent similarity and the input coordinates themselves to construct the point cloud and determine correspondence, replacing the coordinate regression done by the decoder. Extensive experiments show that our construction scheme leads to a performance boost in comparison to recent state-of-the-art correspondence methods. Our code is publicly available\footnote{\url{https://github.com/dvirginz/DPC} \\ \textsuperscript{*}Equal contribution}.
\end{abstract}

%% file: 01_introduction.tex
\section{Introduction} \label{sec:introduction}
With the rise of the availability and vast deployment of 3D data sensors in various fields, 3D computer vision research has thrived. A core problem in 3D vision is shape correspondence. That is, finding a dense mapping from one shape to another. This information opens the door to a variety of applications, including non-rigid human body alignment, articulated motion transfer, face swapping, and more.

\input{figures/teaser/teaser.tex}

%The shape correspondence problem has been thoroughly investigated for 3D mesh data. A mesh is a set of vertices and faces, describing the surface of the 3D shape. \itai{A prominent approach for such data is functional mapping, which aligns the spectral eigenbasis of the shapes~\cite{litany2017deep, ginzburg2020cyclic, roufosse2019unsupervised, donati2020deep}. This approach has been has been proven to be extremely successful. However, it requires the computation of the eigenbasis for each shape~\cite{rustamov2007laplace}, which is a resource-demanding and time-consuming pre-processing step. Moreover, the spectral decomposition uses the mesh connectivity information. This information is often absent in a real-world scenario, when the data originates from a 3D sensing device and contains only the point coordinates.}

The shape correspondence problem has been thoroughly investigated for 3D mesh data~\cite{vankaick2011asurvey, biasotti2016recent}. Recent works have taken a spectral approach by computing the functional mapping between the projected features of the shapes onto their Laplace-Beltrami Operator (LBO) eigenbasis~\cite{litany2017deep, ginzburg2020cyclic, roufosse2019unsupervised, donati2020deep}. Functional mapping methods rely on the global geometric shape structure and learn a transformation between the eigendecomposition of source and target shapes, which is then translated to a point-to-point correspondence.

%The shape correspondence problem has been thoroughly investigated for 3D mesh data. A 3D mesh is a set of vertices and faces, describing the surface of the 3D shape. Researchers have proposed spatial methods for discriminative vertex representation and matching, based on vertex properties and face-based connectivity~\cite{masci2015geodesic, boscaini2016anisotropic, monti2017geometric}. The functional mapping approaches rely on the global geometric shape structure and learn a transformation between the eigendecomposition of a source and target shapes, which is then translated to a point-to-point correspondence.

Spectral matching techniques for 3D meshes have been proven to be extremely successful. However, the computation of the eigenbasis for each shape is a resource-demanding and time-consuming pre-processing step. Even more consequential, the use of such methods in deep learning pipelines is unstable, making them impractical in many cases, as demonstrated in the literature~\cite{ginzburg2020dual} and encountered again in this paper. Moreover, the computation of the LBO basis functions~\cite{rustamov2007laplace} requires the connectivity information between the mesh vertices. Such information is often absent in a real-world scenario when the data originates from a 3D sensing device and contains only the point coordinate information.

%\dvir{\footnote{GeoFMNet and SURFMNet fail to converge on SMAL and TOSCA due to numerical instabilities in functional mapping derivation step}} % too early, we have not talked on the datasets yet.

Recently, correspondence methods for point cloud data have been proposed~\cite{groueix20183dcoded, deprelle2019learning, zeng2020corrnet3d}. Point-based techniques typically employ an encoder-decoder architecture, where the decoder is used as a proxy for determining point assignments. For example, 3D-CODED~\cite{groueix20183dcoded} deformed a given template shape to reconstruct different point clouds. The decoder regressed each point cloud's coordinates and the correspondence was set according to proximity to the given points in the template.

The use of a decoder burdens the learning process of point cloud matching. Since the decoder performs a regression operation, it demands a large amount of training data. Additionally, the decoder is adapted to the distribution of the training shapes and limits the generalization capability of point-based methods. Up-to-date, neither shape correspondence method has overcome the shortcoming in generalization power while presenting real-time performance.

%Up-to-date, neither method for shape correspondence has overcome the shortcoming in generalization power while presenting real-time performance.

%Recently, correspondence methods for point cloud data have been proposed~\cite{groueix20183dcoded, deprelle2019learning, zeng2020corrnet3d}, alleviating the drawbacks of their mesh-based counterparts. However, point-based techniques typically perform shape reconstruction by a training a decoder, as a proxy for determining point assignments. The decoder burdens the learning process, demands a large amount of training data, and limits the generalization capability across datasets. Up-to-date, neither method was able to overcome the lack of generalization power, while presenting real-time performance.

%Up-to-date, a technique that overcomes the lack of generalization power, while presenting real-time performance, has not been proposed.

%\open{Until today, no work was able to overcome the lack of generalization power, while presenting state-of-the-art results and real-time performance.}

In this work, we present a novel technique for real-time dense point cloud correspondence that learns discriminative point embedding without a decoder. Instead, for each source point, we approximate the corresponding point by a weighted average of the target points, where the weights are determined according to a local neighborhood in the learned feature space. We call this operation \textit{cross-construction}, as it uses the existing target points themselves and the latent affinity to construct the target shape rather than regress the points with a decoder. Similarly, we use a self-construction operation for each shape to constrain the learned features to be spatially smooth, thus, reducing outlier matches and improving the correspondence accuracy. Our method does not require matching points supervision, as we use the given input point clouds in the optimization of the construction results.

%In this work, we present a novel technique that befits from both worlds. We use a deep learning-based feature extractor that consumes point clouds directly and does not require any pre-processing, nor point connectivity knowledge. Then, we measure the similarity between points in the two shapes in the learned feature space. Finally, we use the affinity measure and the point cloud data itself to cross-construct the shapes by each other, avoiding a post-processing decoding step of shape coordinate regression. Our cross-construction operation enables us to determine per-point source to target correspondence, without any correspondence supervision.

Extensive evaluations on major shape matching benchmarks of human and non-human figures demonstrate the advantages of our method. Compared to previous works, we show substantial performance gains while using a fraction of the training data. To summarize, our key contributions are as follows:
\begin{itemize}[noitemsep, nolistsep]
    \item We propose a novel unsupervised real-time approach for dense point cloud correspondence\footnote{38 shape pairs per second, where each shape contains 1024 points.}. It employs point similarity in a learned feature space and structured construction modules, rather than a point regression by a decoder network, as done in existing works;
    \item Our method can be trained on small datasets and exhibits compelling generalization capability;
    \item We surpass the performance of existing methods by a large margin for human and animal datasets.
\end{itemize}

%% file: figures/teaser/teaser.tex
\begin{figure}[tb!]
\begin{center}
\begin{tabular}{c c}
\includegraphics[width=0.40\linewidth]{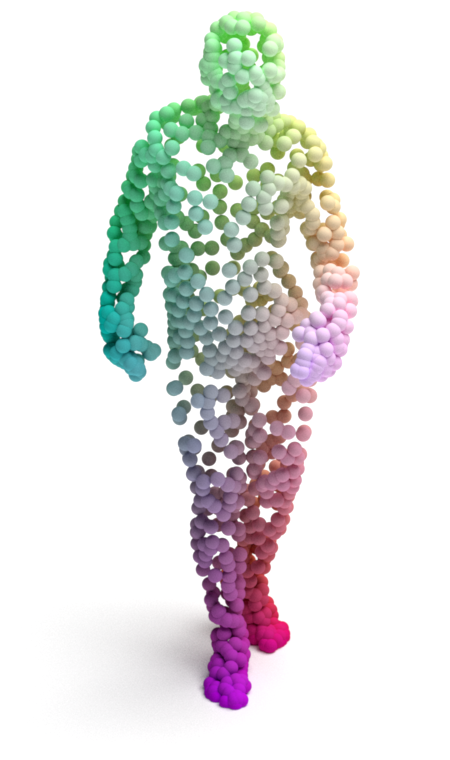} &
\includegraphics[width=0.40\linewidth]{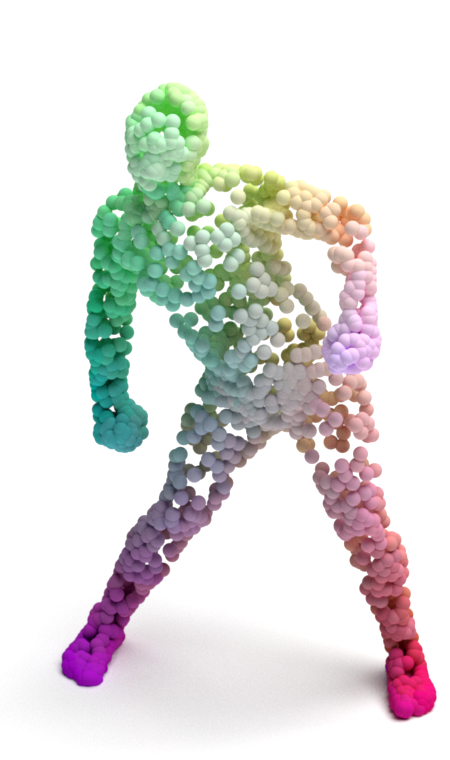} \\
\includegraphics[width=0.40\linewidth]{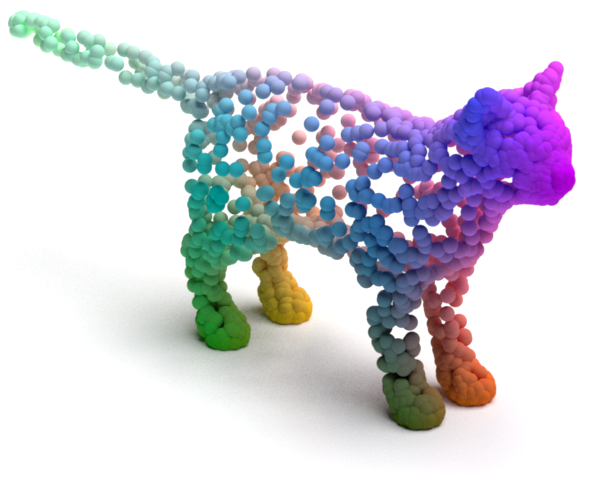} &
\includegraphics[width=0.40\linewidth]{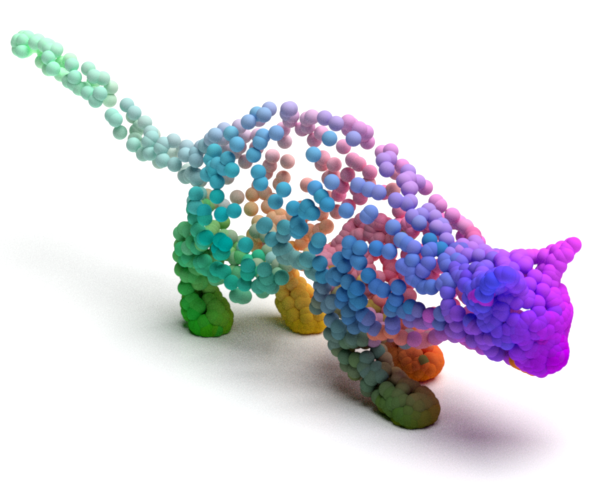} \\
\whitetext{aa}Reference shape & \whitetext{aa}Our result \\
\end{tabular}
\caption{\textbf{Correspondence by DPC.} Our method learns a fine-grained point-to-point mapping between point clouds without matching supervision. It is versatile and can be applied to various non-rigid shapes, including humans and animals. The correspondence is visualized by transferring colors from the left to the right shape according to our resulting matches.}
\label{fig:teaser}
\end{center}
\end{figure}

%We present DPC, an unsupervised method for dense point cloud correspondence. Our methods leverages learned self and cross similarity between points to produce a point-to-point map between the shapes  (color coded).

%% file: 02_related_work.tex
\section{Related Work} \label{sec:related_work}

\noindent \textbf{Feature extraction} \quad Powerful feature learning techniques have been proposed for 3D point clouds~\cite{qi2017pointnet, thomas2019kpconv, wu2019pointconv, wang2019dynamic}. PointNet~\cite{qi2017pointnet} and PointNet++~\cite{qi2017pointnetpp} pioneered an MLP-based learning approach directly on raw point clouds. DGCNN~\cite{wang2019dynamic} extended these works and proposed an edge-convolution operator, along with dynamic feature aggregation. As DGCNN has been proven to be a discriminative feature extractor, we use it in our work as the backbone network for deep point embedding.

% \noindent \textbf{Feature extraction for 3D data} \quad Feature extraction is a fundamental building block in computer vision algorithms. Its aim is to compute a discriminative representation for a data point, where other properties, such as smooth representation across neighboring points, may also be considered. Various hand-crafted descriptors have been proposed to 3D mesh data~\cite{sun2009aconcise, tombari2010unique, aubry2011the}, representing the local geometry of a vertex based on the mesh connectivity and other vertices properties, such as curvature and normal orientation. In recent years, the research focus has been shifted towards a learned feature representation, computed by deep neural networks. In such a framework, optimal feature embedding is learned for the correspondence task~\cite{masci2015geodesic, boscaini2016anisotropic, monti2017geometric}.

% Powerful learning techniques have also been proposed for 3D point clouds~\cite{qi2017pointnet, thomas2019kpconv, wu2019pointconv, wang2019dynamic}, for which the underlying point connectivity is unknown. PointNet~\cite{qi2017pointnet} and PointNet++~\cite{qi2017pointnetpp} pioneered an MLP-based learning approach directly on raw point clouds, eluding pre-processing operations, such as multi-view projection or voxelization. DGCNN~\cite{wang2019dynamic} extended these works and proposed an edge-convolution operator, along with dynamic feature aggregation. As DGCNN has been proven to be a discriminative feature extractor, we use it in our work as the backbone for deep point embedding.

\medskip

\noindent \textbf{Shape correspondence} \quad A substantial body of research has been dedicated to the task of dense shape correspondence~\cite{vankaick2011asurvey, biasotti2016recent}. The goal of this task is to find a point-to-point mapping between a pair of shapes. A prominent approach for mesh-represented shapes is based on functional maps, in which a linear operator is optimized for spectral shape bases alignment~\cite{ovsjanikov2012functional, roufosse2019unsupervised, ginzburg2020cyclic, donati2020deep, eisenberger2020deep}. One advantage of this technique is the structured correspondence prediction using the learned functional map. However, the calculation of the spectral basis comes at the cost of high computational demand and pre-processing time and requires mesh connectivity information. Lately, Marin \etal~\cite{marin2020correspondence} suggested a spectral matching approach for point clouds. They circumvented the LBO-based eigendecomposition by employing a neural network to learn basis functions for the shapes. Then, these functions were used in a functional-maps framework to derive point correspondences.

%They circumvented the LBO-based eigendecomposition by employing a neural network to learn basis functions for the shapes and used these functions in a functional-maps framework to derive point correspondences.

%surface representation of the shape to be known.

In another line of work, a spatial approach has been taken. This kind of works can operate directly on raw 3D point clouds without the need for a costly pre-processing step. Recently, Groueix \etal~\cite{groueix20183dcoded} and Deprelle \etal~\cite{deprelle2019learning} proposed to extract point correspondences via a learned deformation of a template shape. Zeng \etal~\cite{zeng2020corrnet3d} presented a deep neural network to learn a matching permutation between two point clouds, guided by the reconstruction of both shapes. These methods rely on a decoder network to regress the shape coordinates for reconstruction. This decoder trammels the learning process, as in addition to point representation, the model also needs to learn weights for the reconstruction part. In contrast, we employ a structured shape construction approach. Rather than regressing the point cloud coordinates, we use its original points and a similarity measure between learned point representations to cross-construct the shapes for determining correspondence.

\medskip

\noindent \textbf{Point cloud construction} \quad Our construction operation relates to works in the literature from other domains, such as point cloud sampling and registration~\cite{lang2020samplenet, wang2019deep, wang2019prnet}. Lang \etal~\cite{lang2020samplenet} learned to sample a point cloud as a combination of points in the complete point set. As opposed to our work, they employed a similarity metric in the raw point space and used sparse point clouds. Wang and Solomon~\cite{wang2019deep, wang2019prnet} used a learned feature embedding to map one point cloud to another other for computing a global rigid transformation between the two point clouds that represent the same rigid shape. Our work differs fundamentally. We seek to find per-point assignments between point clouds that represent non-rigid shapes.

%% file: 03_method.tex
\section{Method}  \label{sec:method}
A point cloud is a set of unordered 3D point coordinates $\Xcal \in \mathbb{R}^{n \times 3}$, where $n$ is the number of points. Given two point clouds $\Xcal, \Ycal \in \mathbb{R}^{n \times 3}$, referred to as source and target, respectively, our goal is to find a mapping $f: \Xcal \rightarrow \Ycal$, such that for each point $x_i \in \Xcal$, we obtain the corresponding point $y_{j^*} \in \Ycal$, where $1 \leq i, j^* \leq n$. %Our approach operates on raw point clouds and does not require point connectivity information. %without pre-processing. %Additionally, we assume that ground-truth matching is unknown, which enables us to work with unannotated data collections.

Our approach operates on raw point clouds and does not require point connectivity information. A diagram of the method is presented in Figure~\ref{fig:system}. We divide the method into three components: deep feature extraction, affinity matrix computation, and cross and self-construction. The first part leverages a deep neural network to learn a high-dimensional point feature embedding. Next, we measure similarity between points in the two point sets based on their learned representation. Finally, we use unsupervised construction modules that drive the learning process to produce discriminative and smooth point feature fields, which are suitable for the dense matching problem.

\input{figures/system/system_pdf.tex}
\input{figures/cross_similarity/cross_similarity_pdf.tex}

\subsection{Per-point Feature Embedding} \label{subsec:feat_emb}
To learn a high-dimensional point representation for correspondence $F_{\Xcal}, F_{\Ycal} \in \mathbb{R}^{n \times c}$, where $c$ is the feature dimension, we use a variant of the DGCNN model~\cite{wang2019dynamic}. DGCNN applies a series of learned convolutions on the difference between the features of a point and its neighbors. The neighbors are set dynamically according to their feature representation for exploiting non-local point information. This architecture has been proven very efficient for high-level tasks, such as point cloud classification and semantic segmentation. In these tasks, there are typically a few tens of classes.

%These tasks typically have a few tens of classes.

However, in our case, the point affiliation is in the order of thousands of candidate corresponding points. Moreover, as the 3D shapes often have co-occurred similar segments, like human arms or animal legs, such a dynamic neighborhood frequently leaks information from one segment to the other, causing symmetry inconsistencies, as discussed previously in the literature~\cite{ginzburg2021dwc}. Thus, we use a static neighborhood graph and describe each point according to its local geometry to increase the granularity of the point feature representation. The complete architecture details are provided in the supplementary material.

\subsection{Similarity Measure} \label{subsec:sim_measure}
The matching between the source and target point clouds is determined according to the similarity between the points' latent embedding (Figure~\ref{fig:cross_similarity}). We measure proximity as the cosine of the angle between their feature vectors:
\begin{equation} \label{eq:similarity}
s_{ij} = \frac{F_{\Xcal}^i \cdot (F_{\Ycal}^j)^T}{||F_{\Xcal}^i||_2 ||F_{\Ycal}^j||_2},
\end{equation}

\noindent where $F_{\Xcal}^i, F_{\Ycal}^j \in \mathbb{R}^{c}$ are the $i$'th and $j$'th rows of $F_{\Xcal}$ and $F_{\Ycal}$, respectively, and $(\cdot)^T$ denotes a transpose operation. $S_{\Xcal\Ycal}$ denotes the affinity matrix between the point sets, where $S_{\Xcal\Ycal}^{ij} = s_{ij}$.

We use the cosine similarity since it inherently includes feature vector normalization and its range is bounded. These properties contribute to the stability of our correspondence learning pipeline.

%We use the cosine similarity due to its beneficial properties. First, it inherently includes feature vector normalization, which regularizes the learning process and enables the network to explore the point latent space. Second, the cosine similarity range is bounded. This property contributes to the numerical stability of our point matching pipeline.

\subsection{Construction Modules} \label{subsec:construct_modules}
Our dense correspondence learning process is unsupervised. To learn the point representation without matching annotations, we use two novel modules: cross-construction and self-construction. The cross-construction operator promotes unique point matches between the shape pair, while the self-construction operator acts as a regularizer and encourages the correspondence map to be smooth.

%--------------------------------------- Cross-construction
\medskip
\noindent \textbf{Cross-construction} \quad The cross-construction module uses the latent proximity between source and target points and the target point coordinates to construct the target shape. The module's operation is depicted in Figure~\ref{fig:cross_construction}. For each source point $x_i \in \Xcal$, we employ a softmax operation and normalize the similarity to its $k_{cc}$ nearest neighbors, to form a distribution function:
\begin{equation} \label{eq:cross_weights}
w_{ij} = \frac{e^{s_{ij}}}{\sum_{l \in \Neu_\Ycal(x_i)} e^{s_{il}}},
\end{equation}

\noindent where $\Neu_\Ycal(x_i)$ is the latent cross-neighborhood, which contains the $k_{cc}$ indices of $x_i$'s latent nearest neighbors in $\Ycal$. Then, we compute an approximate matching point $\hat{y}_{x_i}$ as:
\begin{equation} \label{eq:soft_cross_corr}
\hat{y}_{x_i} = \sum_{j \in \Neu_\Ycal(x_i)}{w_{ij} y_j}.
\end{equation}

\noindent The cross-construction of the target $\Ycal$ by the source $\Xcal$ is denoted as $\widehat{\Ycal}_\Xcal \in \mathbb{R}^{n \times 3}$, where $\widehat{\Ycal}_\Xcal^i = \hat{y}_{x_i}$.

The point $\hat{y}_{x_i}$ approximates the target point corresponding to $x_i$. In order to encourage a unique matching between the source and target points, we would like that each point in $\widehat{\Ycal}_\Xcal$ will have a close point in $\Ycal$ and vice versa. Thus, we minimize the Chamfer Distance~\cite{achlioptas2018learning} between the two. The Chamfer Distance between two point clouds $\Pcal$ and $\Qcal$ is given by:
\begin{equation} \label{eq:chamfer_dist}
\begin{split}
C&D(\Pcal, \Qcal) = \\
  &\frac{1}{|\Pcal|}\sum_{p \in \Pcal}{\min_{q \in \Qcal}||p - q||_2^2} +
   \frac{1}{|\Qcal|}\sum_{q \in \Qcal}{\min_{p \in \Pcal}||q - p||_2^2}.
\end{split}
\end{equation}

\noindent Thus, we define a target cross-construction loss as:
\begin{equation} \label{eq:cross_loss_target}
\Lcal_{cc}(\Ycal, \widehat{\Ycal}_\Xcal) = CD(\Ycal, \widehat{\Ycal}_\Xcal).
\end{equation}

%NOTE: the cycle-consistent point mapping claim is not accurate when the number of points is different in the source and target. Also, we do not show consistency analysis in the paper. Thus, this claim should not be used!!!
%\noindent To encourage a cycle-consistent point mapping, we also define a loss for the cross-construction of the source by the target, where the source and target switch roles:

\input{figures/cross_construction/cross_construction_pdf.tex}

Since the target can also be matched to the source, we can exploit this mapping direction as another training example for our model. Thus, we define a loss for the cross-construction of the source by the target, where the source and the target switch roles:
\begin{equation} \label{eq:cross_loss_source}
\Lcal_{cc}(\Xcal, \widehat{\Xcal}_\Ycal) = CD(\Xcal, \widehat{\Xcal}_\Ycal),
\end{equation}

\noindent and use it in the overall training objective function of our model.

%--------------------------------------- Self-construction
\medskip
\noindent \textbf{Self-construction} \quad A desired property of a dense correspondence field is smoothness~\cite{vestner2017product, ginzburg2020dual, zeng2020corrnet3d}. That is, geometrically close points should have similar latent representations. To promote this property, we propose the self-construction module. Like in the cross-construction operation, a point $x_i \in \Xcal$ is approximated by its $k_{sc}$ neighboring latent points:
\begin{equation} \label{eq:soft_self_corr}
\hat{x}_i = \sum_{l \in \Neu_\Xcal(x_i)}{w_{il} x_l},
\end{equation}

%That is, neighboring points in the source should correspond to neighboring points in the target. In our case, it implies that neighbor points should have similar feature vectors.

%\dvir{I believe the statement is not clear enough, as it sounds like there is a connection between source and target in this module, maybe something like:}  \dvir{geometrically close points should have close latent representation, while dissimilar latent representation to geometrically far points}

\noindent where $\Neu_\Xcal(x_i)$ is the latent self-neighborhood of $x_i$ in $\Xcal \setminus x_i$, $w_{il}$ are computed as in Equation~\ref{eq:cross_weights} with $\Neu_\Xcal(x_i)$ instead of $\Neu_\Ycal(x_i)$, and the similarity in Equation~\ref{eq:similarity} is measured between $F_\Xcal^i$ and $F_\Xcal^l$ instead of $F_\Xcal^i$ and $F_\Ycal^j$. This process is repeated for the target shape as well. Finally, we would like to minimize the self-construction loss terms, defined as:
\begin{equation} \label{eq:self_loss_source}
\begin{split}
\Lcal&_{sc}(\Xcal, \widehat{\Xcal}_\Xcal) = CD(\Xcal, \widehat{\Xcal}_\Xcal) \\
\Lcal&_{sc}(\Ycal, \widehat{\Ycal}_\Ycal) = CD(\Ycal, \widehat{\Ycal}_\Ycal).
\end{split}
\end{equation}

\noindent When these loss terms are minimized, a point is approximated by its neighbors in the shape. Thus, they have similar features and the smoothness property is achieved, which in turn reduces outlier matches and improves the correspondence accuracy.

We note that our self-construction loss is analogous to the Laplace-Beltrami operator (LBO)~\cite{rustamov2007laplace}, which computes the difference between a point and the average of its neighbors. However, while in the LBO operator the neighbors are set according to proximity in the raw Euclidean space, in our loss term, the neighbors and their weights are determined by the affinity measure in a learned feature space.

\subsection{Training Objective and Matching Inference} \label{subsec:obj_and_infer}
In addition to the construction modules, we regularize the mapping of neighboring points such that close points in the source will correspond to close points in the target. To this end, we use the following mapping loss:
\begin{equation} \label{eq:neigh_loss_target}
\Lcal_{m}(\Xcal, \widehat{\Ycal}_\Xcal) = \frac{1}{n k_m} \sum_i \sum_{l \in N_\Xcal(x_i)} {v_{il} ||\hat{y}_{x_i} - \hat{y}_{x_l}||_2^2},
\end{equation}

\noindent where $N_\Xcal(x_i)$ is the Euclidean neighborhood of $x_i$ in $\Xcal \setminus x_i$ of size $k_m$, $\hat{y}_{x_i}$ is defined in Equation~\ref{eq:soft_cross_corr}, $v_{il} = e^{-||x_i - x_l||_2^2/\alpha}$ weight the loss elements according to the proximity of the source points, and $\alpha$ is a hyperparameter. We note that this loss is complementary to the self-construction loss. The mapping loss further assists in reducing outlier matches and improves the mapping accuracy. Similar to the other loss terms, we also define this loss for the mapping from the target to the source, namely, $\Lcal_{m}(\Ycal, \widehat{\Xcal}_\Ycal)$. An analysis of the importance of the different loss terms is discussed in an ablation study in sub-section~\ref{subsec:ablation_study}. The overall objective function of our point correspondence learning scheme is:
\begin{equation} \label{eq:total_loss}
\begin{split}
\Lcal_{total} = \lambda_{cc}&(\Lcal_{cc}(\Ycal, \widehat{\Ycal}_\Xcal) + \Lcal_{cc}(\Xcal, \widehat{\Xcal}_\Ycal)) + \\ \lambda_{sc} &(\Lcal_{sc}(\Xcal, \widehat{\Xcal}_\Xcal) + \Lcal_{sc}(\Ycal, \widehat{\Ycal}_\Ycal))+ \\ \lambda_{m} &(\Lcal_{m}(\Xcal, \widehat{\Ycal}_\Xcal) + \Lcal_{m}(\Ycal, \widehat{\Xcal}_\Ycal)),
\end{split}
\end{equation}

\noindent where $\lambda_{cc}$, $\lambda_{sc}$, and $\lambda_{m}$ are hyperparameters, balancing the contribution of the different loss terms.

At inference time, we set the target point with the highest weight in the latent cross-neighborhood of $x_{i}$ to be its corresponding point. That is:
\begin{equation} \label{eq:hard_proj}
f(x_i) = y_{j^*}, \; j^* = \argmax_{j \in \Neu_\Ycal(x_i)} w_{ij}.
\end{equation}

\noindent This selection rule can be viewed as a local classification of the source point to the most similar point in its latent cross-neighborhood in the target. The point $y_{j^*}$ is also the closest latent neighbor to $x_{i}$ from $\Ycal$, as:
\begin{equation} \label{eq:closest_cross_neighbor}
j^* = \argmax_{j \in \Neu_\Ycal(x_i)} w_{ij} = \argmax_{j \in \Neu_\Ycal(x_i)} s_{ij} = \argmax_{j} s_{ij},
\end{equation}

%\noindent This selection rule can be viewed as the classification of the source point to its most similar target point. Indeed, $y_{j^*}$ is closest neighbor in the feature space to $x_{i}$ from the target shape, as:

\noindent where the first transition is due to Equation~\ref{eq:cross_weights} and the second is from the latent cross-neighborhood $\Neu_\Ycal(x_i)$ definition.

%% file: figures/system/system_pdf.tex
\begin{figure*}[tb!]
\begin{center}
\includegraphics[width=\linewidth]{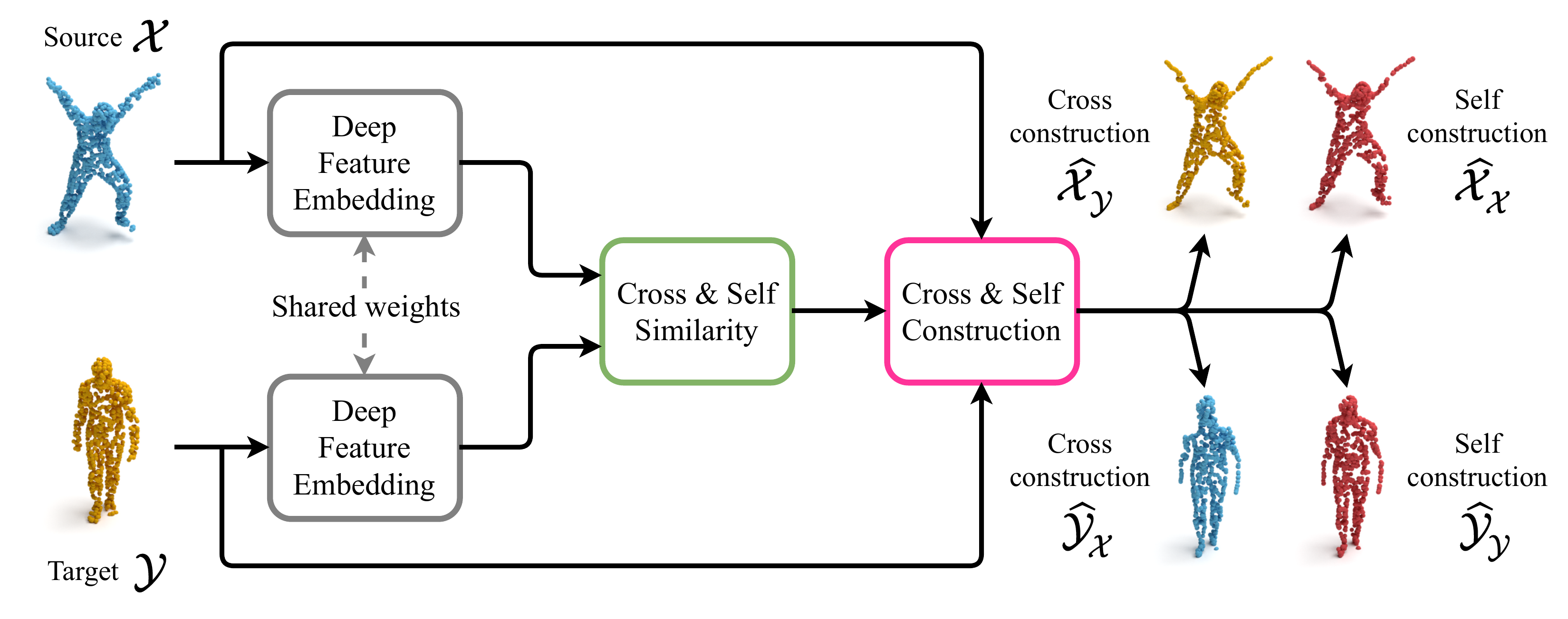}
\caption{{\bfseries The proposed method.} Our method operates directly on point clouds and does not require point connectivity information nor ground-truth correspondence annotations. The method consists of three parts: a per-point learned feature embedding, cross and self-similarity measure in the feature space, and cross and self-construction modules. The construction modules guide the learning process to produce discriminative yet smooth point representation, suitable for a point-wise mapping between the shapes.}
\vspace{-10pt}
\label{fig:system}
\end{center}
\end{figure*}

%% file: figures/cross_similarity/cross_similarity_pdf.tex
\begin{figure}[tb!]
\begin{center}
\includegraphics[width=0.9\linewidth]{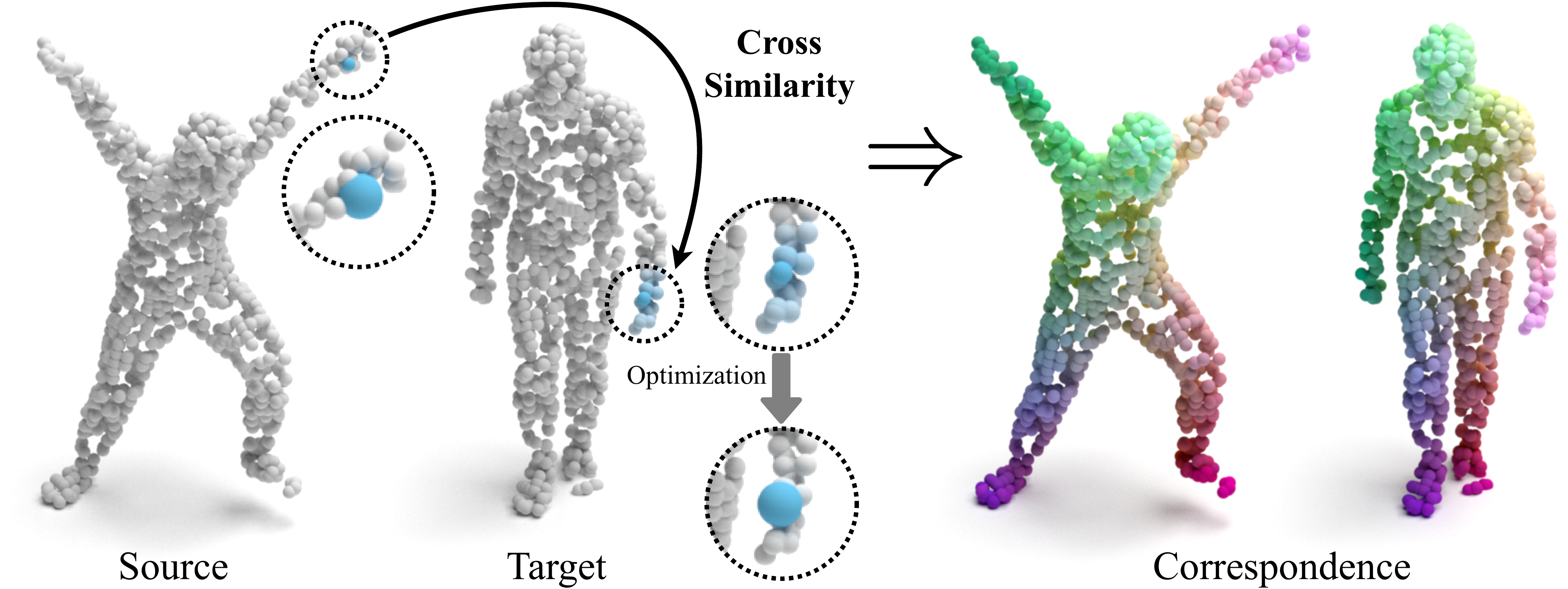}
\caption{\textbf{Cross-similarity illustration}. On the left shape pair, we mark an example source point (enlarged) and its closest points in the feature space in the other shape. The features are optimized via our construction modules to obtain a similar and unique target point embedding (enlarged), which results in a dense correspondence map between the point clouds. The mapping is color-coded in the shape pair on the right.}
\label{fig:cross_similarity}
\end{center}
\end{figure}

%We mark an example point in the source shape and its closest target points in the feature space (left). The features are optimized via our construction modules to have unique point matches, which results in a dense correspondence map between the point clouds (right).

%The latent neighbors are located at the corresponding region to the example point, where the most similar neighbor is considered as the matching point.

%% file: figures/cross_construction/cross_construction_pdf.tex
\begin{figure}[tb!]
\begin{center}
\includegraphics[width=\columnwidth]{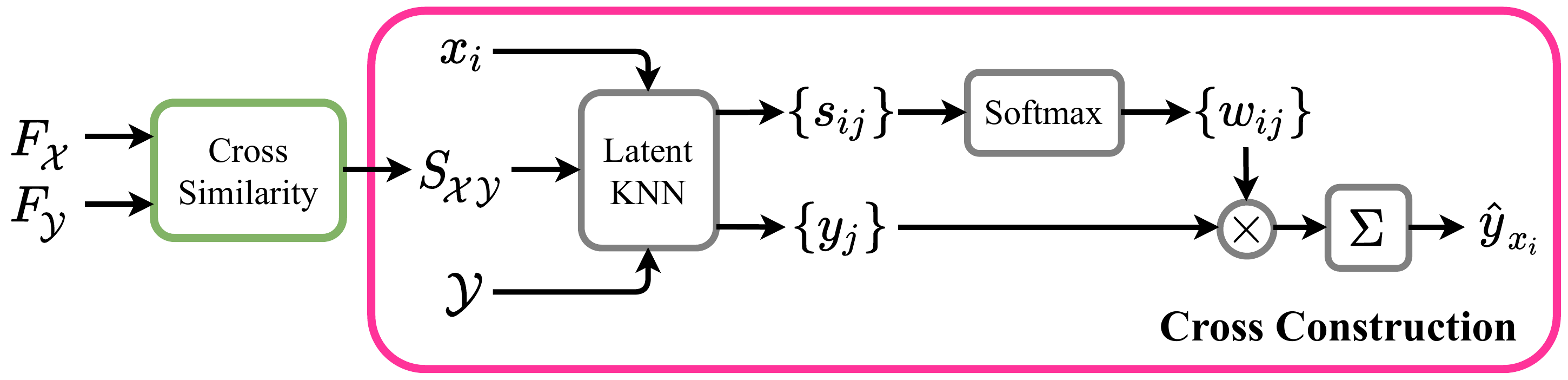}
\caption{{\bfseries The cross-construction module.} The module leverages similarity in the latent feature space to find a neighborhood of a source point $x_i$ in the target point cloud $\Ycal$. The neighbor points are weighted according to their affinity to the source point to construct the point $\hat{y}_{x_i}$, which approximates the correspondence of $x_i$ in $\Ycal$.}
\label{fig:cross_construction}
\end{center}
\end{figure}

%% file: 04_results.tex
\section{Results} \label{sec:resutls}
%Overview of the section (which experiments are conducted and why).
In this section, we present the results of our DPC model on several well-established datasets and compare the model's performance to recent state-of-the-art techniques for shape matching. Additionally, we evaluate the run time of the different methods and discuss ablation experiments that validate the design choices in our method.

%Additionally, we conduct a thorough ablation study and validate the design choices in our method.

\input{figures/comparison_shrec/comparison_shrec_pdf.tex}
\input{figures/surreal_on_shrec/surreal_on_shrec_pdf.tex}

%-------------------------------------------------------------------------
\subsection{Experimental Setup} \label{subsec:experimental_setup}
\noindent \textbf{Datasets} \quad To demonstrate the flexibility of our method, we evaluate it on both human and non-human datasets. For human figures, we use the SURREAL training dataset of 3D-CODED~\cite{groueix20183dcoded}, consisting of 230000 training shapes. This is a synthetic dataset generated from the parametric human model SMPL~\cite{loper2015smpl}. During training, we select shapes from the dataset at random and use them as training pairs. For testing, we follow the evaluation protocol of previous works~\cite{groueix20183dcoded, deprelle2019learning, zeng2020corrnet3d} and use the challenging SHREC'19~\cite{melzi2019matching} dataset. This dataset contains 44 real human scans, paired into 430 annotated test examples.

% consisting of 230K and 200 shapes for training and testing, respectively.

%These datasets are widely used by prior methods, and we adopt them in our work as well.

For non-human shapes, SMAL~\cite{zuffi20173dmenagerie} and TOSCA~\cite{bronstein2008numerical} datasets are adopted for training and evaluation, respectively. SMAL is a parametric model for animals: cat, dog, cow, horse, and hippo. We use the model to create 2000 examples in various poses for each animal type and obtain a train set of 10000 shapes in total. For training, we randomly select shape pairs from within the same animal category. The TOSCA collection includes 80 objects of animals and humans. Different from SMAL, it is not a model-based dataset and also includes animal species other than those in SMAL. We consider the 41 animal figures from TOSCA and pair shapes from the same category to form a test set of 286 examples.

\input{tables/surreal_and_shrec/acc_and_error.tex}

%For non-human shapes, SMAL~\cite{zuffi20173dmenagerie} and TOSCA~\cite{bronstein2008numerical} datasets are adopted for training and evaluation, respectively. SMAL is a parametric model for animals: cat, dog, cow, horse, and hippo.

%The TOSCA collection includes 80 objects and animals and humans. Different from SMAL, this is not a model-based dataset. From Tosca we take the 41 animal figures, which also include different species that those of SMAL, and pair them to a test set of size 286.

%It was used for creating synthetic dataset of several animals: cat, dog and horse, in various poses.

The number of points in the shapes substantially varies from one dataset to another. Thus, to have a shared baseline, we randomly sample $n = 1024$ points from each shape to create point clouds for training and evaluation, as done in CorrNet3D work~\cite{zeng2020corrnet3d}. In the supplementary, we also report results for higher point cloud resolutions.

\medskip

\noindent \textbf{Evaluation metrics} \quad A common evaluation metric for shape correspondence is the average geodesic error~\cite{litany2017deep, donati2020deep}, which requires knowing the point adjacency matrix. However, as we are in the realm of point clouds, we assume that this information is unavailable and adopt a Euclidean-based measure instead~\cite{zeng2020corrnet3d}. For a pair of source and target shapes $(\Xcal, \Ycal)$, the correspondence error is defined as:
\begin{equation} \label{eq:corr_err}
err = \frac{1}{n}\sum_{x_i \in \Xcal}{||f(x_i) - y_{gt}||_2},
\end{equation}

\noindent where $y_{gt} \in \Ycal$ is the ground truth corresponding point to $x_i$\footnote{Ground truth data is only used during testing DPC, not during its training.}. Additionally, we measure the correspondence accuracy, defined as:
\begin{equation} \label{eq:corr_acc}
acc(\epsilon) = \frac{1}{n}\sum_{x_i \in \Xcal}{\mathbbm{1}{(||f(x_i) - y_{gt}||_2} < \epsilon d}),
\end{equation}

\noindent where $\mathbbm{1}{(\cdot)}$ is the indicator function, $d$ is the maximal Euclidean distance between points in $\Ycal$, and $\epsilon \in [0, 1]$ is an error tolerance. We note that these evaluation metrics are analogous to the ones used for mesh-represented shapes, with the geodesic distance replaced by the Euclidean distance.

\medskip

\noindent \textbf{Implementation details} \quad We implement our method in PyTorch~\cite{paszke2017automatic} and adapt the open-source DGCNN~\cite{wang2019dynamic} implementation for our feature extractor module. For the cross and self-construction operations and the mapping loss, we use a neighborhood size of $k_{cc} = k_{sc} = k_m = 10$. $\lambda_{cc}$, $\lambda_{sc}$, and $\lambda_m$ in Equation~\ref{eq:total_loss} are set to $1$, $10$, and $1$, respectively. Additional implementation details are given in the supplementary material. 

\medskip

\noindent \textbf{Baseline methods} \quad We contrast our method with the recent state-of-the-art point-based matching techniques Diff-FMaps~\cite{marin2020correspondence}, 3D-CODED~\cite{groueix20183dcoded}, and Elementary Structures work~\cite{deprelle2019learning}. These methods require ground-truth matching supervision. Additionally, we compare with the most recent CorrNet3D work~\cite{zeng2020corrnet3d} that learns point cloud correspondence in an unsupervised manner. For the completeness of the discussion, we also include in our evaluation the unsupervised SURFMNet~\cite{roufosse2019unsupervised} and the supervised GeoFMNet~\cite{donati2020deep} methods for meshes. For all the examined baselines, training and evaluation are done using their publicly available official source code.

%\dvir{I think we can drop this sentece, we've already stated that} These works takes a spectral approach based on functional maps, and have shown outstanding corresponding performance.

%Training and evaluation for all the examined baselines is done using their official open-source code, where we optimize their performance to the examined test cases detailed hereinafter.

%We contrast our method with the popular 3D-CODED~\cite{groueix20183dcoded} and Elementary Structures~\cite{deprelle2019learning} works. These works use ground-truth matching supervision. We also compare to the most recent unsupervised approach CorrNet3D~\cite{zeng2020corrnet3d}.

%For a comprehensive comparison, we also include in our evaluation the unsupervised SURFMNet~\cite{roufosse2019unsupervised} and the supervised GeoFMNet~\cite{donati2020deep} methods for meshes. For the baseline methods, we use their official open-source code, and optimize their performance to the examined test cases detailed hereinafter.

%\noindent \textbf{Training and evaluation protocol} For training on SURREAL, we randomly pair the available shapes in the training set. The evaluation is done on the official 430 pairs in SHREC test set. We also examine the capability of our model to be trained on the small SHREC 

%Datasets (train/test), evaluation metrics (including justification, \ie, why we selected these datasets and metrics), implementation details.

\input{tables/time_analisys/time_analisys.tex}

%-------------------------------------------------------------------------
\subsection{Evaluation on Human Datasets} \label{subsec:human_results}
\noindent \textbf{Cross-dataset generalization} \quad Figure~\ref{fig:surreal_on_shrec} presents the correspondence accuracy (Equation~\ref{eq:corr_acc}) for point-based methods trained on SURREAL and evaluated on the SHREC'19 test set. For a fair comparison, we report the results without any post-processing and include methods that operate on point sets and do not require the additional connectivity information, nor the intense eigendecomposition pre-processing step, as the spectral approaches demand. As seen in the figure, our method outperforms the competing approaches by a large margin. For example, at a 5\% error tolerance, it achieves 50\% accuracy, an improvement of 23\% over CorrNet3D~\cite{zeng2020corrnet3d}. Visual examples in Figure~\ref{fig:comp_shrec} demonstrate the improved accuracy of our method compared to the previous ones.

%Figure~\ref{fig:surreal_on_shrec} presents the point correspondence performance for several point-based methods trained on SURREAL~\cite{varol2020learning, groueix20183dcoded} and evaluated on the official SHREC'19~\cite{melzi2019matching} test set. To show that our method can be trained on a small amount of data, we randomly select 2K examples from SURREAL train set, while for the other baseline methods we use all the available 230K shapes.

%Description of the datasets (how many shapes, synthetitc or real shapes), description of the compared methods and why we compare to them.

%\noindent \bluetext{Figure: graph of correspondence accuracy percentage vs. error tolerance (in euclidean distance percentage). Train on SURREAL 2k and test on SHREC. Compare to CorrNet3D, DeepGFM, 3D-CODED, SURFMNet (optional), Elementary Structures (optional).}

Diff-FMaps~\cite{marin2020correspondence} has learned basis functions from the synthetic SURREAL data, which are less suitable for aligning the different SHREC'19 shapes. The other compared methods employ a learned reconstruction module as a proxy for point matches. This module requires a large amount of training data and limits the generalization capability to the unseen test dataset. In contrast, we take a spatial approach and exclude the point regression of the decoder. Instead, we replace it with our structured construction modules and concentrate on point feature learning for the correspondence task. It enables our method to be trained on only a small fraction (1\%) of 2000 SURREAL shapes, while the other methods utilize all the available 230000 instances.

%\input{tables/ablation_study/ablation_study.tex}

%our method learns point similarity and uses the given points of the shape itself to construct it. We note that while the other methods are trained on 230K shapes, we used only 2K shapes for training, two orders magnitude less.

\medskip

\noindent \textbf{Training on a small dataset} \quad To further demonstrate the ability of our method to learn discriminative point representations from a small amount of data, we train it on random shape pairs from the 44 human instances in SHREC'19. These pairs do not have matching annotations and are suitable only for unsupervised techniques. Table~\ref{tbl:surreal_and_shrec} reports the correspondence accuracy at 1\% tolerance, which represents a near-perfect hit, and the average correspondence error (Equation~\ref{eq:corr_err}) for training either on SURREAL or SHREC shapes and testing on the official 430 SHREC'19 pairs.

%which is important for dense correspondence applications, such as texture transfer. We use SURREAL and SHREC datasets, and evaluate the matching in intra- and inter-dataset settings. For a comprehensive comparison, in addition to the alternative methods in the experiment before, we also compare to the recent state-of-the-art mesh-based methods SURFMNet~\cite{roufosse2019unsupervised} and GeoFMNet~\cite{donati2020deep}.

%In Table~\ref{tbl:surreal_and_shrec} we focus on the correspondence accuracy at an error tolerance of 1\%. This working point represents a near-perfect matching, which is important for dense correspondence applications, such as texture transfer. We use SURREAL and SHREC datasets, and evaluate the matching in intra- and inter-dataset settings. For a comprehensive comparison, in addition to the alternative methods in the experiment before, we also compare to the recent state-of-the-art mesh-based methods SURFMNet~\cite{roufosse2019unsupervised} and GeoFMNet~\cite{donati2020deep}.

The spectral methods SURFMNet~\cite{roufosse2019unsupervised} and GeoFMNet~\cite{donati2020deep} show an outstanding result for the correspondence error. However, they are less accurate at 1\% tolerance. We believe that this is due to the projection of the vertex feature maps on the highly smooth spectral functions, which reduces the overall error but compromises the near-perfect accuracy. We also show in sub-section~\ref{subsec:time_analysis} that the spectral methods are approximately $100 \times$ slower compared to DPC in terms of total inference run-time, making them impractical for real-time usage. Among point-based approaches, we achieve the best results on both the accuracy and error measures. Notably, our method reaches a comparable performance when trained either on SURREAL or SHREC, where the latter contained only 44 shapes, two orders of magnitude less than the former (which includes 2000 shapes).

%GeoFMNet improves over the baseline point methods. However, it projects learned shape descriptors on a truncated spectral basis of the shape, which results in an over-smooth correspondence prediction. In contrast, we use proximity between the learned point embedding directly and show considerable performance gains compared to GeoFMNet. Additional, as our method is unsupervised, it can be trained on the SHREC dataset, which do not have ground truth annotations for each pair of shapes. Although this is a very small dataset that contains only 44 shapes, our method achieves performance comparable for the inter-dataset setting, where it was trained on 2000 SURREAL shapes.

\input{figures/comparison_tosca/comparison_tosca.tex}
\input{figures/smal_on_tosca/smal_on_tosca_pdf.tex}

%-------------------------------------------------------------------------
\subsection{Time Analysis} \label{subsec:time_analysis}
We evaluate the average processing time of different methods for computing the correspondence between a pair of shapes from the SHREC'19 test set. The measurements were done on an NVIDIA RTX 2080Ti GPU. Table~\ref{tbl:time_analysis} summarizes the results. The spectral methods~\cite{roufosse2019unsupervised, donati2020deep} require the time-consuming LBO-based spectral decomposition of each shape, which results in a long overall time duration. In contrast, DPC operates directly on raw input point clouds without any pre-processing and runs faster than the other spectral and point-based alternatives. Its inference time is only 26.3 milliseconds, providing a real-time processing rate of 38 point cloud pairs per second. To sum up, our method offers a sweet spot of strong generalization capabilities, along with real-time performance.

%-------------------------------------------------------------------------
\subsection{Evaluation on Non-human Shapes} \label{subsec:non_human_results}
We demonstrate the flexibility of our method by applying it to the dense alignment of animal point sets. Similar to the evaluation for human figures, we examine its and the other methods' generalization power by training on the model-based SMAL dataset and testing on the diverse animal objects from the TOSCA set. Figure~\ref{fig:comp_tosca} shows visualizations and Figure~\ref{fig:smal_on_tosca} depicts the correspondence accuracy results.

%Similar to the evaluation on human figures, we train the compared methods on SMAL dataset and evaluate the performance on the TOSCA test pairs. Figure~\ref{fig:smal_on_tosca} depicts the results. 

Both 3D-CODED~\cite{groueix20183dcoded} and Elementary Structures~\cite{deprelle2019learning} rely on the deformation of a template shape to the source and target point clouds for deducing the correspondence map between them. In the SMAL dataset, this template takes the form of a standing cat. However, the TOSCA set includes substantially different poses and shapes, such as a howling wolf. Thus, these methods struggle to generalize to this test case.

%\open{Furthermore, while the parametric space of SMAL is quite limited, TOSCA shape poses are diverse, with complex postures (\eg, a cat standing on its rear legs) found in the data.}

CorrNet3D~\cite{zeng2020corrnet3d}, on the other hand, does not depend on a template shape and improves over 3D-CODED and Elementary Structures. Still, it includes a decoder module that is fitted to the characteristics of the SMAL data and compromises CorrNet3D generalization capability to TOSCA's animal objects. Our method neither uses a template nor a decoder component. Instead, it learns robust local point descriptors, which enables it to operate on shapes out of the training distribution. As visualized in Figure~\ref{fig:comp_tosca} and quantified in Figure~\ref{fig:smal_on_tosca}, our DPC consistently surpasses the performance of the alternatives for point cloud matching.

Lastly, we refer the reader to the supplementary for an evaluation on SMAL and TOSCA, where we present our accuracy at 1\% tolerance and the average correspondence error for these datasets as well. Since SURFMNet~\cite{roufosse2019unsupervised} and GeoFMNet~\cite{donati2020deep} are LBO-based architectures, they fail to be applied to the SMAL and TOSCA sets since these methods are numerically unstable under non-watertight or topology-intersected meshes, as present in the datasets.

%CorrNet3D~\cite{zeng2020corrnet3d}, on the other hand, does not depend on a template shape and improves over 3D-CODED and Elementary Structures. Still, it includes a decoder module that is fitted to the characteristics of the SMAL data and compromises CorrNet3D generalization capability to animal shapes different that those in the training data. our method does not depend on a template and do not have a decoder component.  Instead, it concentrates on learning local discriminative point features, which enables it to operate on shapes out of the training distribution. As seen in Figure 8, it surpasses CorrNet3D accuracy across all the examined error tolerance range.

%-------------------------------------------------------------------------
\subsection{Ablation Study} \label{subsec:ablation_study}
In the supplementary, we present a thorough ablation study verifying the design choices in our DPC model. From the ablation study, we recognize that the local neighborhood for cross-construction contributes the most to the method's performance. While considering all target points for mapping a source point is a common approach in previous works~\cite{wang2019deep, zeng2020corrnet3d, eisenberger2021neuromorph}, we find it less effective, as it reduces our correspondence accuracy by 15.5\%. Instead, for each source point, we consider a local latent neighborhood in the target shape. It focuses the model on exploring only relevant candidates for matching and eases the learning process. Additionally, the ablation study highlights the importance of our self-construction module, which regularizes the learned point representation and is crucial for an accurate correspondence result. Without it, the performance drops by 14.3\%. For further details, please see the supplementary.

%In the supplementary, we present an ablation study that verifies the design choices in our DPC model. From the ablation study, we recognize that the local neighborhood for cross-construction contributes the most to the method's performance. While considering all target points for mapping of a source point is a common approach in previous works~\cite{wang2019deep, zeng2020corrnet3d, eisenberger2021neuromorph}, we find it less effective in our case. Instead, for each source point, we consider a local latent neighborhood in the target shape. It focuses the model on exploring only relevant candidates for matching and eases the learning process. Additionally, the ablation study highlights the importance of our self-construction module, which regularizes the learned point representations and is crucial for an accurate correspondence result. Further details are given in the \open{supplementary}.

%In such cases, the cross-construction error is high, and the shape matching is compromised.

%% file: figures/comparison_shrec/comparison_shrec_pdf.tex
\begin{figure*}[tb!]
\begin{center}
\begin{tabular}{c c c c c c}
\multicolumn{6}{c}{\includegraphics[width=0.96\linewidth]{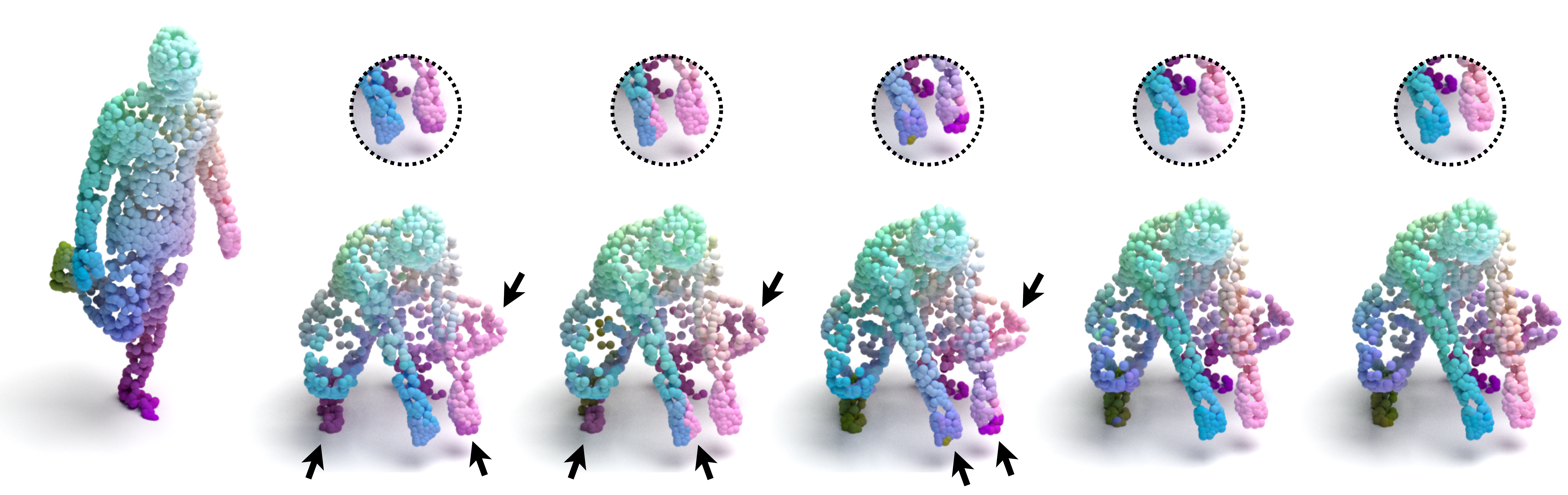}} \\
\whitetext{a}Reference target & \whitetext{}3D-CODED~\cite{groueix20183dcoded} & \whitetext{aa}Elementary~\cite{deprelle2019learning} & \whitetext{a}CorrNet3D~\cite{zeng2020corrnet3d} & \whitetext{aaa}DPC (ours) & \whitetext{aa}Ground-truth \\
\end{tabular}
\end{center}
\caption{\textbf{Visual comparison for human shapes for a SHREC'19 test pair.} The training was done on the SURREAL dataset. Previous methods suffer from correspondence errors, such as matching the knee to the hand or mixing between the limbs (marked with arrows and zoomed-in). In contrast, our method produces an accurate result, which is close to the ground truth correspondence map (color-coded).}
\label{fig:comp_shrec}
\end{figure*}

%Previous methods suffer from correspondence errors, such as matching the leg to hand or mixing between the hands (marked with arrows and zoomed-in).

%% file: figures/surreal_on_shrec/surreal_on_shrec_pdf.tex
\begin{figure}[tb!]
\begin{center}
\includegraphics[width=\columnwidth]{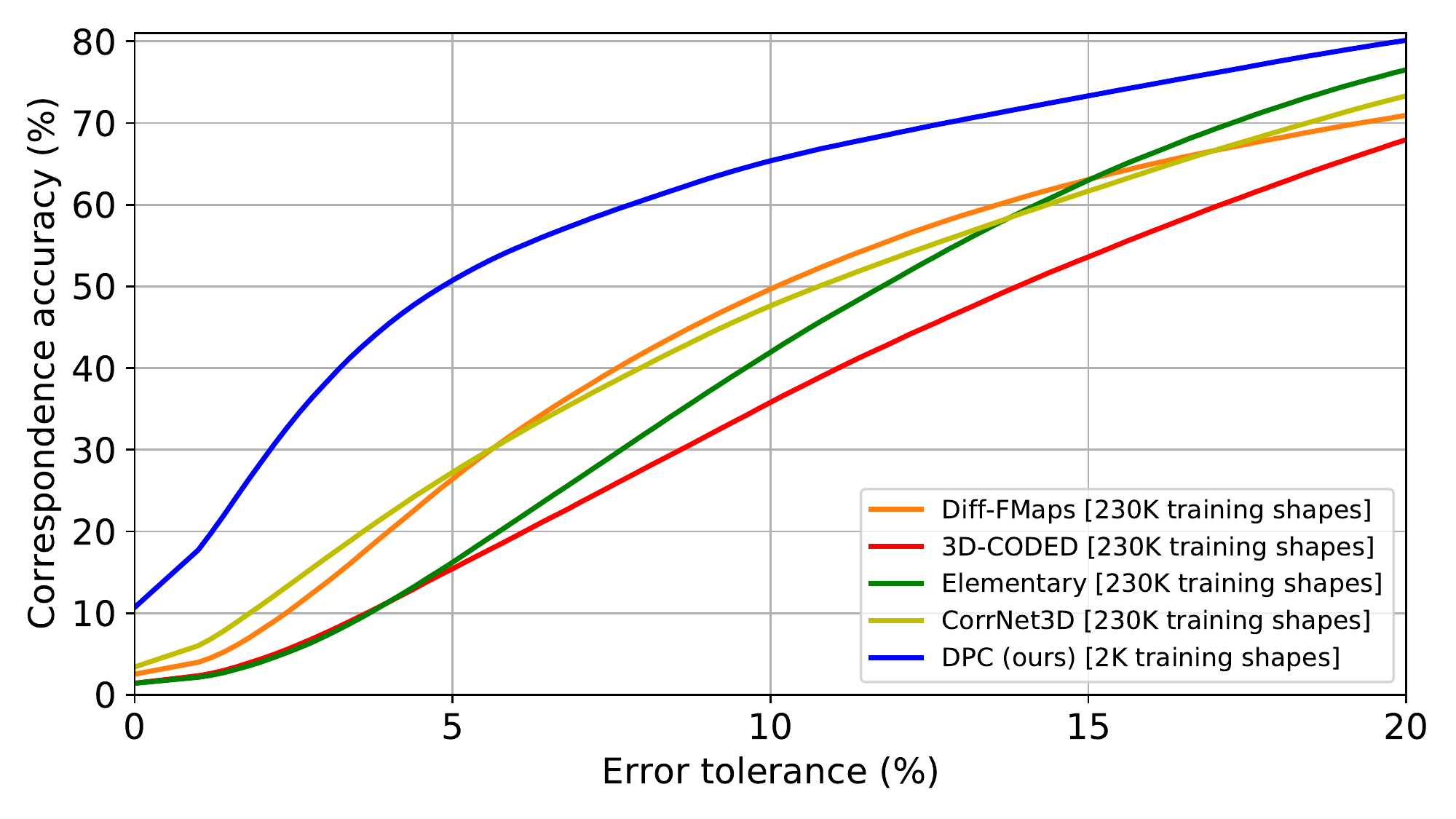}
\caption{{\bfseries Correspondence accuracy for human point clouds.} The methods were trained on the SURREAL dataset and evaluated on the official SHREC'19 test pairs. The number of training shapes (in thousands) is stated in the squared brackets. Our method achieves a substantial performance gain compared to the alternatives while being trained on much less training data.}
\vspace{-5pt}
\label{fig:surreal_on_shrec}
\end{center}
\end{figure}

%% file: tables/surreal_and_shrec/acc_and_error.tex
\begin{table}[tb!]
\small
\centering
\begin{tabular}{@{ } l c c c c c @{}}
\toprule
       & & \multicolumn{2}{c}{\textbf{SURREAL/}} & \multicolumn{2}{c}{\textbf{SHREC/}} \\
       & & \multicolumn{2}{c}{\textbf{SHREC}}    & \multicolumn{2}{c}{\textbf{SHREC}}  \\
       \cmidrule(lr){3-4} \cmidrule(lr){5-6}
Method & Input  & \textit{acc} $\uparrow$ & \textit{err} $\downarrow$ & \textit{acc} $\uparrow$ & \textit{err} $\downarrow$\whitetext{a}\: \\
\midrule
\graytxt{SURFMNet~\cite{roufosse2019unsupervised}} & \graytxt{LBO} & \graytxt{4.3\%} & \graytxt{0.3} & \graytxt{5.9\%} &  \graytxt{0.2} \\
\graytxt{GeoFMNet~\cite{donati2020deep}} & \graytxt{LBO} & \graytxt{8.2\%} & \graytxt{0.2} & \graytxt{*} & \graytxt{*} \\
\midrule
Diff-FMaps~\cite{marin2020correspondence} & Point & 4.0\% & 7.1 & * & *        \\
3D-CODED~\cite{groueix20183dcoded}        & Point & 2.1\% & 8.1 & * & *        \\
Elementary~\cite{deprelle2019learning}    & Point & 2.3\% & 7.6 & * & *        \\
CorrNet3D~\cite{zeng2020corrnet3d}        & Point & 6.0\% & 6.9 & 0.4\% & 33.8 \\
DPC (ours) & Point & \textbf{17.7\%} & \textbf{6.1} & \textbf{15.3\%} & \textbf{5.6} \\
\bottomrule
\end{tabular}
\vspace{0.1cm}
\caption{\textbf{Accuracy and error.} We evaluate the accuracy at 1\% tolerance (\textit{acc}, in percentage) and the average correspondence error (\textit{err}, in centimeters) for two train/test settings of human datasets. Higher accuracy and lower error reflect a better result. Spectral methods on meshes are grayed out to emphasize the difference in their operation requirements compared to ours. We achieve better results than the other point-based methods.}
\label{tbl:surreal_and_shrec}
\end{table}

%Among the point cloud-based methods, we achieve the best results for both metrics.

%The best results for point cloud-based methods are marked in bold.

%We evaluate the accuracy at 1\% tolerance and the average correspondence error for two train/test settings of Human datasets. The latter setting do not have ground-truth correspondence annotations, and thus, supervised methods cannot be applied (denoted by *). Spectral methods on meshes are grayed out to emphasize the difference in their operation requirements.

%% file: tables/time_analisys/time_analisys.tex
\begin{table}[tb!]
\small
\centering
\begin{tabular}{@{ } l c c c @{ }}
\toprule
Method & Pre-process & Inference & Total \\
\midrule
\graytxt{SURFMNet~\cite{roufosse2019unsupervised}} & \graytxt{1593} & \graytxt{163} & \graytxt{1756} \\
\graytxt{GeoFMNet~\cite{donati2020deep}}           & \graytxt{1997} & \graytxt{215} & \graytxt{2212} \\
\midrule
Diff-FMaps~\cite{marin2020correspondence}     & 0 & 121.7  & 121.7 \\
3D-CODED~\cite{groueix20183dcoded}            & 0 & 32.1  & 32.1 \\
Elementary~\cite{deprelle2019learning}        & 0 & 35.3  & 35.3 \\
CorrNet3D~\cite{zeng2020corrnet3d}            & 0 & 175.4 & 175.4 \\
DPC (ours)                                    & 0 & \textbf{26.3} & \textbf{26.3} \\
\bottomrule
\end{tabular}
\vspace{0.1cm}
\caption{\textbf{Processing time comparison.} We report the time requirements (in milliseconds) for different methods for the correspondence computation. Our method has the best time performance, which is about $100\times$ faster compared to the spectral methods~\cite{roufosse2019unsupervised, donati2020deep}.}
\label{tbl:time_analysis}
\end{table}

%% file: figures/comparison_tosca/comparison_tosca.tex
\begin{figure*}[tb!]
\begin{center}
\begin{tabular}{c c c c c c}
% % cat
% \includegraphics[width=0.14\linewidth]{figures/comparison_tosca/dpc/s_07_corr_bac.png} & % target gt
% \includegraphics[width=0.14\linewidth]{figures/comparison_tosca/3dcoded/t_07_corr_bac.png} & % 3D-CODED
% \includegraphics[width=0.14\linewidth]{figures/comparison_tosca/elementary/t_07_corr_bac.png} & % Elementary Structures
% \includegraphics[width=0.14\linewidth]{figures/comparison_tosca/corrnet3d/t_07_corr_bac.png} & % CorrNet3D
% \includegraphics[width=0.14\linewidth]{figures/comparison_tosca/dpc/t_07_corr_bac.png} & % DPC (ours)
% \includegraphics[width=0.14\linewidth]{figures/comparison_tosca/dpc/t_07_gt_bac.png} \\ % source gt

% wolf
\includegraphics[width=0.14\linewidth]{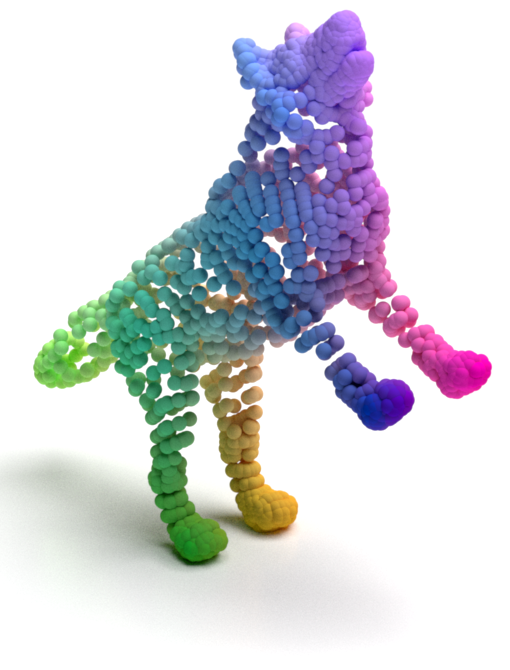} & % target gt
\includegraphics[width=0.14\linewidth]{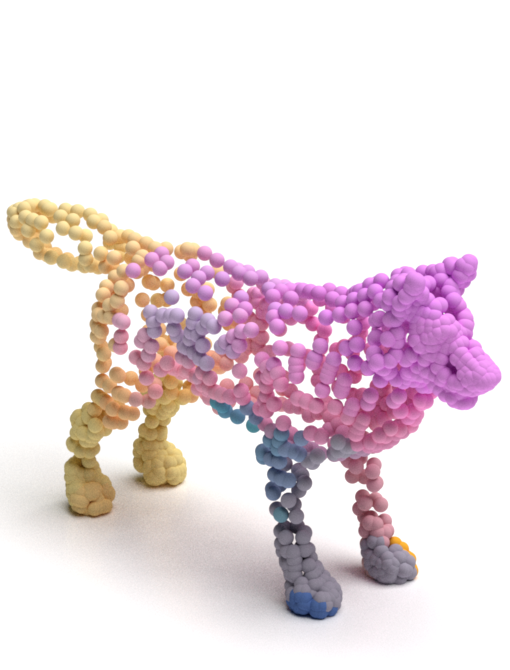} & % 3D-CODED
\includegraphics[width=0.14\linewidth]{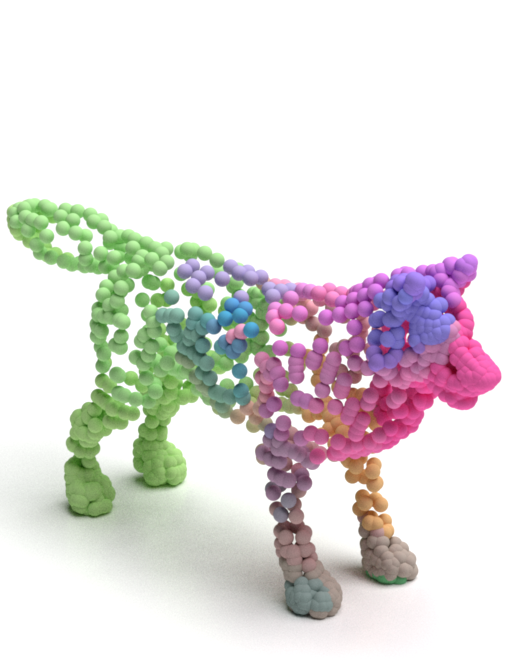} & % Elementary Structures
\includegraphics[width=0.14\linewidth]{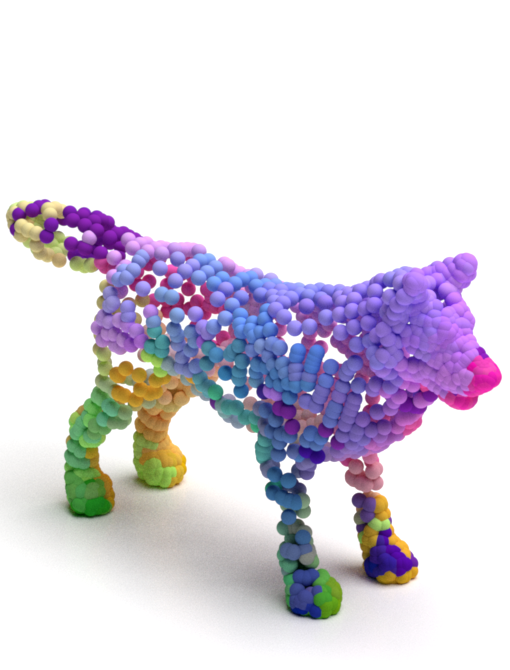} & % CorrNet3D
\includegraphics[width=0.14\linewidth]{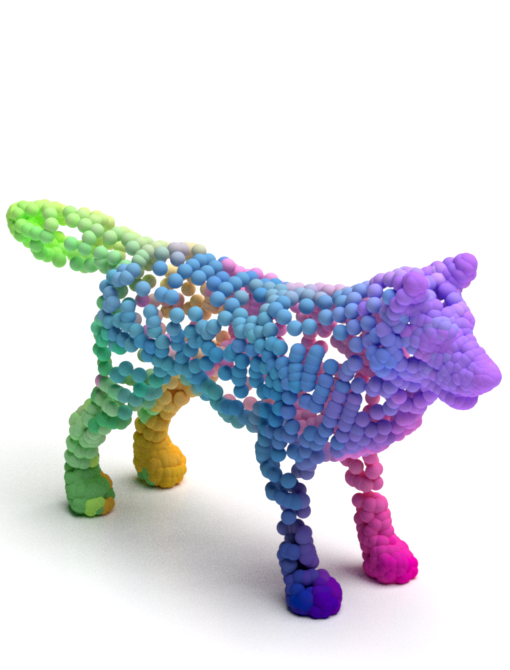} & % DPC (ours)
\includegraphics[width=0.14\linewidth]{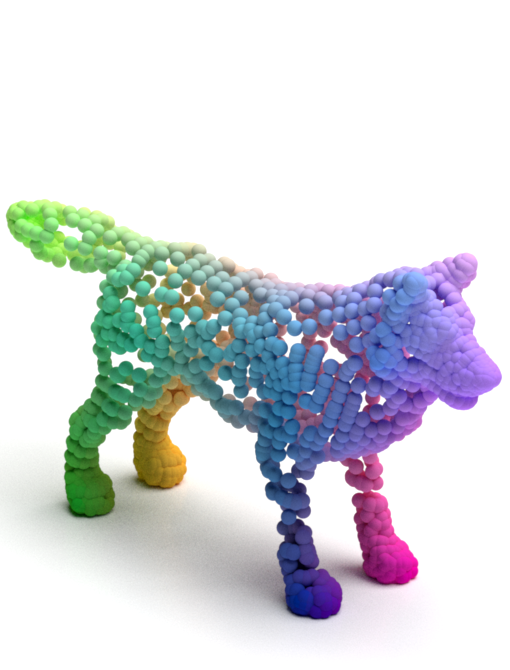} \\ % source gt

% % dog
% \includegraphics[width=0.14\linewidth]{figures/comparison_tosca/dpc/t_142_corr_fwd.png} & % target gt
% \includegraphics[width=0.14\linewidth]{figures/comparison_tosca/3dcoded/s_142_corr_fwd.png} & % 3D-CODED
% \includegraphics[width=0.14\linewidth]{figures/comparison_tosca/elementary/s_142_corr_fwd.png} & % Elementary Structures
% \includegraphics[width=0.14\linewidth]{figures/comparison_tosca/corrnet3d/s_142_corr_fwd.png} & % CorrNet3D
% \includegraphics[width=0.14\linewidth]{figures/comparison_tosca/dpc/s_142_corr_fwd.png} & % DPC (ours)
% \includegraphics[width=0.14\linewidth]{figures/comparison_tosca/dpc/s_142_gt_fwd.png} \\ % source gt

Reference target & 3D-CODED~\cite{groueix20183dcoded} & Elementary~\cite{deprelle2019learning} & CorrNet3D~\cite{zeng2020corrnet3d} & DPC (ours) & Ground-truth \\
\end{tabular}
\caption{\textbf{Visual comparison for animal shapes from the TOSCA test set.} The training was done on the synthetic SMAL dataset. 3D-CODED and Elementary Structures produce a patchy result. CorrNet3D's output is noisy and contains outlier matches. In contrast, our method produces a smooth and accurate alignment between the animal point clouds (color-coded).}
\vspace{-15pt}
\label{fig:comp_tosca}
\end{center}
\end{figure*}

%% file: figures/smal_on_tosca/smal_on_tosca_pdf.tex
\begin{figure}[tb!]
\begin{center}
\includegraphics[width=\columnwidth]{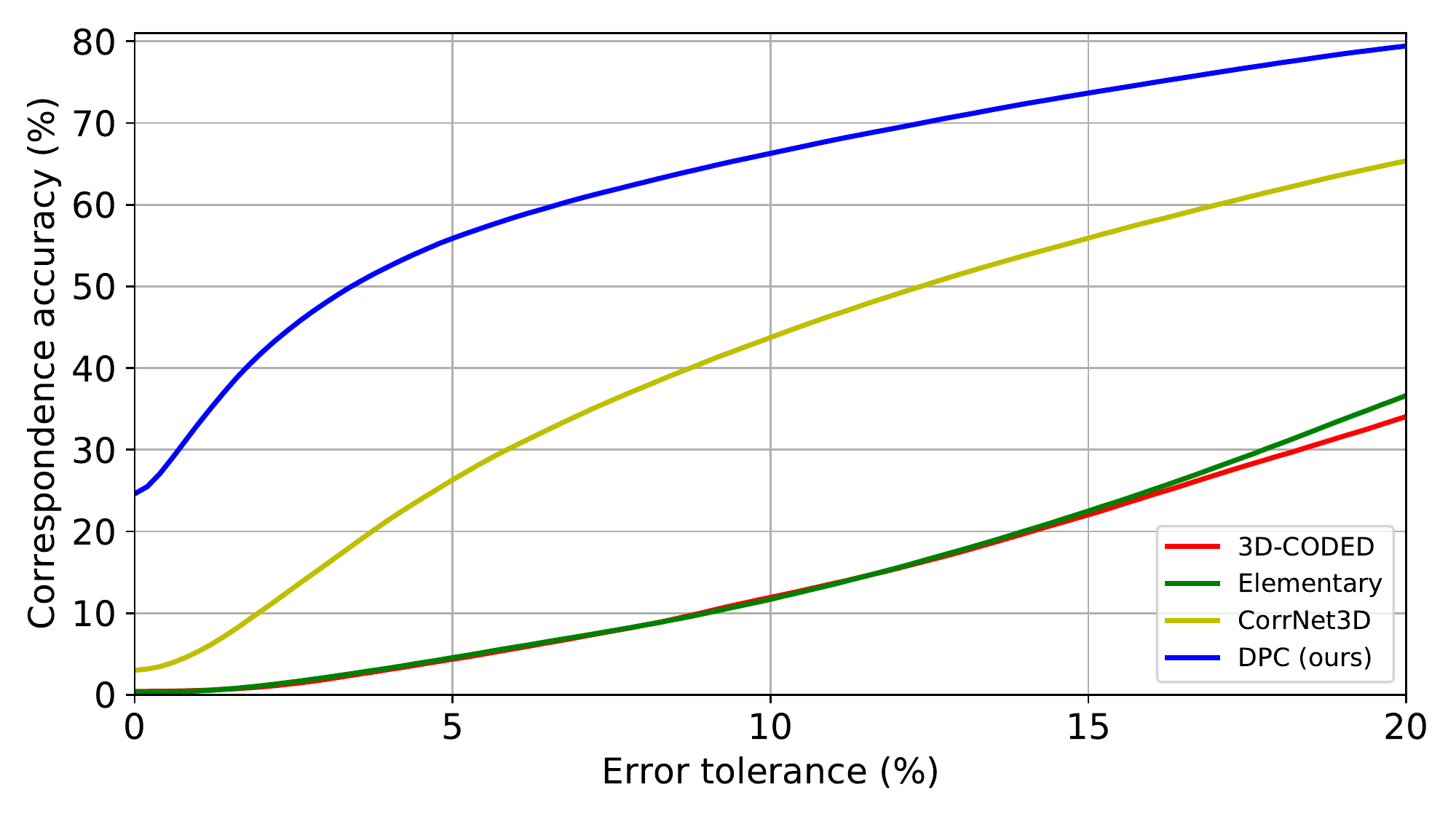}
\caption{{\bfseries Correspondence accuracy for non-humans.} The compared works were trained on the SMAL dataset and tested on animal shapes from the TOSCA benchmark. Our DPC model outperforms the other methods by a large margin across all the error tolerance range.}
\vspace{-15pt}
\label{fig:smal_on_tosca}
\end{center}
\end{figure}

%Our method surpasses previous works by a large margin.

%% file: 05_conclusions.tex
\section{Conclusions} \label{sec:conclusions}
In this paper, we presented a novel unsupervised method for real-time dense correspondence, which operates directly on point clouds without connectivity information. We leverage similarity in a learned embedding space, together with cross and self-construction modules, to compute point-to-point matches without ground truth supervision. Different from previous approaches for point sets, we do not rely on a template shape nor a decoder for point regression but rather concentrate the model on learning discriminative point features for matching. It enables our method to be trained on small datasets while having a compelling generalization ability to other test benchmarks. Evaluation on well-established human and animal datasets showed that our approach surpasses the performance of recent state-of-the-art works on point cloud correspondence by a large margin.

\medskip
\noindent \textbf{Acknowledgments} \quad This work was partly funded by ISF grant number 1549/19 and the Zimin institute for Engineering solutions advancing better lives.

%In this paper we presented a novel unsupervised method for dense point cloud correspondence.  It operates directly on point clouds without connectivity knowledge and does not requires ground truth matching information.  First, our method learns per-point deed feature representation.  Then,it  leverages  similarity  in  the  embedding  space,  between points withing and between point clouds, for self-construction of each shape and cross-constructions of the shapes by each other

%Different from previous approaches that are based on point cloud, we do not rely on a template shape, nor on a decoder for point regression, but rather concentrate on learning discriminative point features for matching. In enables our method to be trained on small datasets, while having a compelling generalization ability to other test benchmarks, surpassing the performance of previous point-based methods by a large margin.

%We evaluated our method on well-established Human as well as Non-human dataset and showed its substantial performance gains in comparison to recent stat-of-the-art works for point cloud correspondence. We also examined several perturbation cases, including noise and partiality, which implies on the robustness and flexibility of our approach.

%Evaluation on well-established Human and Animal datasets showed our substantial performance gains in comparison to recent stat-of-the-art techniques for point cloud correspondence. 

%% file: supplementary/supplementary.tex
\clearpage
\appendix
\section*{Supplementary Material}
In the following sections, we provide more information regarding our point cloud correspondence method. Sections~\ref{sec:human_supp} and \ref{sec:animals_supp} present additional results for human and animal shapes, respectively. An ablation study is reported in Section~\ref{sec:ablation_study_supp}. In Section~\ref{sec:experimental_settings_supp}, we detail experimental settings, including network architecture and optimization parameters of DPC. %Our code is anonymously available on GitHub at \href{https://anonymous.4open.science/r/DeepPC-200F}{https://anonymous.4open.science/r/DeepPC-200F}.

%Finally, in Section~\ref{sec:experimental_settings_supp}, we detail experimental settings, including the network architecture and optimization parameters of DPC. Our code is anonymously available on GitHub at \href{https://anonymous.4open.science/r/DeepPC-200F}{this https URL}.

%-------------------------------------------------------------------------
%%%%%%%%% Additional Human
\input{supplementary/a_additional_human.tex}

%-------------------------------------------------------------------------
%%%%%%%%% Additional Animal
\input{supplementary/b_additional_animal.tex}

%-------------------------------------------------------------------------
%%%%%%%%% Ablation Study
\input{supplementary/c_ablation_study.tex}

%-------------------------------------------------------------------------
%%%%%%%%% Experimental Settings
\input{supplementary/d_experimental_settings.tex}

%% file: supplementary/a_additional_human.tex
\section{Additional Results for Human Datasets} \label{sec:human_supp}

%-------------------------------------------------------------------------
\subsection{Robustness Evaluations} \label{subsec:robustness_evaluations_supp}
We evaluated the ability of our model to infer correspondence for point clouds with different resolutions than the training point clouds. We applied the model that was trained on SURREAL point clouds with 1024 points to SHREC'19 point sets, randomly sampled with a higher number of points in the source and target shapes. The results are quantified in Figure~\ref{fig:point_resolution_supp}. As the figure shows, our method can successfully operate on a point set with $4 \times$ higher resolution, with a small performance drop. 

To further test the robustness of DPC to the point resolution, we applied the model to source and target point clouds that differ in their point number. The result in Figure~\ref{fig:resolution_robustness_supp} demonstrates that our model can gracefully infer the correspondence in this case as well. These experiments suggest that DPC has learned unique and robust point descriptors that vary smoothly between neighboring points, as intended by our constructions modules.

Additionally, we applied our model for matching noisy point clouds and show an example result in Figure~\ref{fig:noise_resilience_supp}. The figure suggests that our method exhibits noise resilience to some extent.

%In this experiment, we apply the model that was trained on SURREAL point clouds with 1024 points, to SHREC'19 point sets, randomly sampled with \textit{different} numbers of points. Figure~\ref{fig:resolution_robustness_supp} shows that the \textit{same} model, without any changes, can gracefully infer the correspondence between point clouds with different resolutions. This property implies that DCP learned unique and robust point descriptors that vary smoothly between neighboring points, as intended by our constructions modules.

% \subsection{Perturbation Robustness} \label{subsec:perturbation_robustness}
% We examine the resistance of our method to several perturbation cases, which are common for point clouds: different number of points in the source and target, noise, and shape partiality. In the first case we use 4096 points for the source and 1024 for the target. In the second, we add Gaussian noise to the source points. Lastly, parts of the shape are removed, by simulating view point occlusion. Figure~\ref{fig:augmentations} visualizes the examined perturbations. As seen in the figure, all these cases are gracefully handled by our method, which indicates its flexibility and robustness.

%-------------------------------------------------------------------------
\subsection{Influence of the Train Set Size} \label{subsec:train_size_supp}
We examined the effect of the train set size on our method's performance by training it on a varying number of point clouds from the SURREAL dataset and evaluating the average correspondence error (equation~\ref{eq:corr_err} in the paper) on the SHREC'19 test set. When DPC was trained on 200 shapes, it resulted in an average error of 6.5 centimeters, an increase of 0.4 centimeters compared to error for using 2000 instances (Table~\ref{tbl:surreal_and_shrec} in the paper). The error reduction for training on 20000 examples was negligible (less than 0.1 centimeters). We conclude that a small training set suffices for our model to converge and achieve an accurate correspondence result.

%We examine the effect of the train set size on our model's performance by training it on a varying number of point clouds from the SURREAL dataset and evaluating the correspondence accuracy on the SHREC'19 test set. When DPC is trained on 200 shapes, the performance reduction is only 2\% compared to using 2000 instances, as done in the paper. The accuracy improvement for training on 20000 examples is negligible, as can be seen in the figure. The same is true for training on all the available 230000 shapes of SURREAL. We conclude that a small training set suffices for our model to converge and achieve accurate correspondence results.

\input{supplementary/figures/point_resolution/point_resolution_pdf.tex}
\input{supplementary/figures/resolution_robustness/resolution_robustness_pdf.tex}
\input{supplementary/figures/noise_resilience/noise_resilience_pdf.tex}

%-------------------------------------------------------------------------
\subsection{Intra-dataset Generalization} \label{subsec:surreal_test_supp}
In the paper, we evaluated the generalization of our model across datasets, when it was trained on SURREAL and tested on SHREC'19. Here, we apply the model to the SURREAL test set to examine the model's generalization capability within the same dataset. The test split of SURREAL includes 200 shapes~\cite{groueix20183dcoded, zeng2020corrnet3d} and we randomly paired them to 1000 test samples. Figure~\ref{fig:surreal_on_surreal_supp} depicts the results. DPC's correspondence result is more accurate than all the compared methods, including the LBO-based approaches \cite{roufosse2019unsupervised, donati2020deep} that use the mesh connectivity and the additional spectral eigenbases information. Notably, our method archives a perfect hit rate of 64\%, an improvement of 44\% over SURFMNet~\cite{roufosse2019unsupervised}. We attribute this result to the descriptive quality of our method, which computes local point representations with a high level of granularity.

%-------------------------------------------------------------------------
\subsection{Additional Visual Results} \label{subsec:additional_vis_supp}
Figure~\ref{fig:additional_comp_shrec_supp} shows additional visual comparison for human shapes. As seen in the figure, our method computes more accurate correspondence results, which are closer to the ground-truth mapping between the point clouds.

% Figure~\ref{fig:additional_comp_shrec_supp} shows additional visual comparison for human shapes. As seen in the figure, our method computes accurate correspondence results, which are close to the ground-truth mapping between the point clouds.

%-------------------------------------------------------------------------
\subsection{Limitations} \label{subsec:limitations_supp}
There are some sources of error for our method. One of them is when body parts are glued to each other. Another is when the body is in a highly unusual position. Examples of such cases are shown in Figure~\ref{fig:failure_cases_supp}. In the first case, the hands are held tight to the body and the matching of the palms and hips is mixed. In the second, the leg is up in the air and is wrongly mapped to the hand. A possible solution is to augment the training data with such shape pairs, which we leave for future work. Notably, the other compared methods also struggle in these challenging cases and fail to align the shapes.

%Our method struggles to compute an accurate correspondence result in some cases. One of them is when the body is in an unusual position. Another is when body parts are glued to each other. We show examples of such failure cases in Figure \redtext{[add figure]}.

%Our method results in high matching errors when the source or target shapes are in a rare position, as demonstrated by Figure \redtext{[add figure]}. A possible solution is to augment the training data with such positions. We leave this direction for a future work. 
%\itai{Need to add failure case visualization (for example: in case of high symmetry, non regular positions. In case of lack of space, the limitations section will be moved to the supplementary.}

%% file: supplementary/figures/point_resolution/point_resolution_pdf.tex
\begin{figure}[tb!]
\begin{center}
\includegraphics[width=\columnwidth]{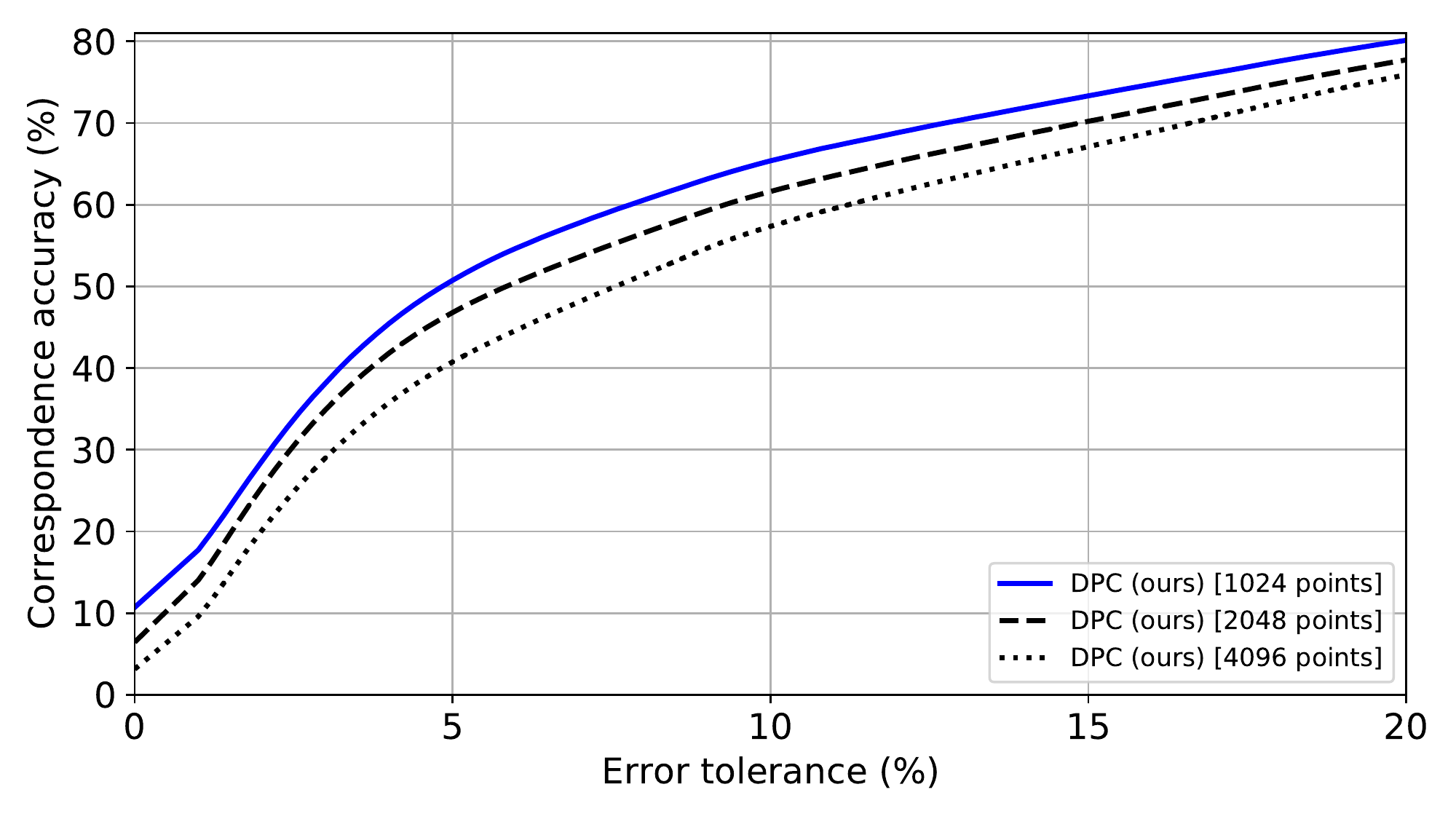}
\caption{{\bfseries Correspondence accuracy for higher point resolutions.} DPC was trained on the SURREAL dataset with a point clouds of 1024 points. The same model, without any modifications, was evaluated on the SHREC'19 test set, with point clouds of higher resolutions (indicated in the squared brackets). Our model can generalize to denser point clouds than those is was trained on, with a mild reduction in performance.}
\label{fig:point_resolution_supp}
\end{center}
\end{figure}

%% file: supplementary/figures/resolution_robustness/resolution_robustness_pdf.tex
\begin{figure}[tb!]
\begin{center}
\includegraphics[width=\linewidth]{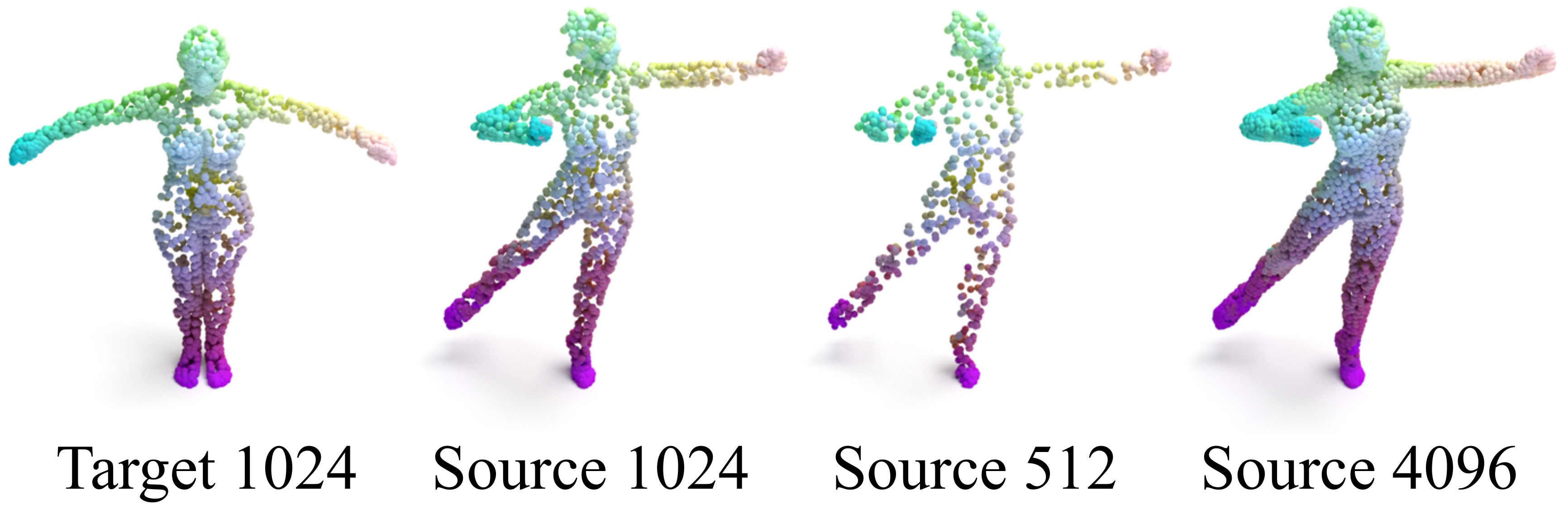} \\
\end{center}
\caption{\textbf{Inference for different number of point.} DPC was trained on SURREAL point clouds with 1024 points. A result for a test pair with 1024 points from the SHREC'19 dataset is shown in the left side. The same model, without any changes, is successfully applied to point clouds with either lower or higher point number (shown on the right side), which implies on model's robustness to the point cloud's resolution. Point matches are indicated by similar colors.}
\label{fig:resolution_robustness_supp}
\end{figure}

%% file: supplementary/figures/noise_resilience/noise_resilience_pdf.tex
\begin{figure}[t!]
\begin{center}
\includegraphics[width=\columnwidth]{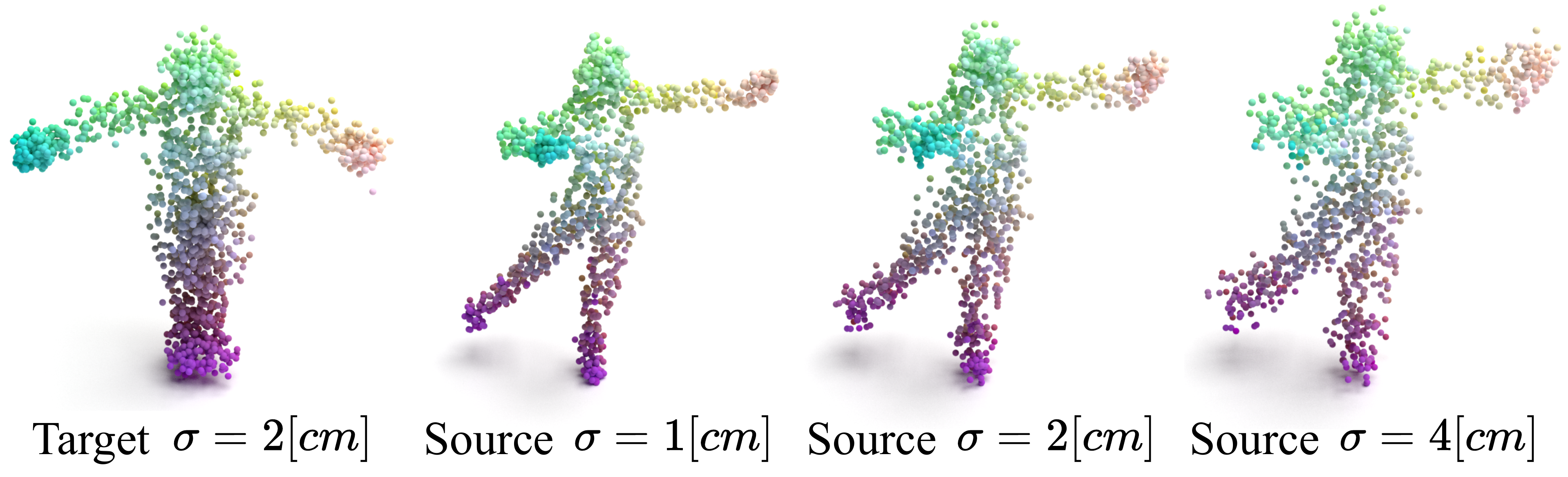}
\caption{{\bfseries Correspondence example for noisy data.} DPC was trained on SURREAL point clouds of 1024 points and evaluated on SHREC'19 point clouds of 1024 points with different levels of added Gaussian noise (indicated by the standard deviation $\sigma$). The point correspondence is color-coded. Our model that was trained on synthetic shapes without noise can gracefully handle the noisy test point clouds.}
\label{fig:noise_resilience_supp}
\end{center}
\end{figure}

%% file: supplementary/b_additional_animal.tex
\section{Additional Results for Animal Shapes} \label{sec:animals_supp}

%-------------------------------------------------------------------------
\subsection{Correspondence Accuracy and Average Error} \label{subsec:acc_err_animal_supp}
In Table~\ref{tbl:smal_and_tosca_supp} we report the correspondence accuracy at 1\% tolerance (Equation~\ref{eq:corr_acc}) and the average matching error (Equation~\ref{eq:corr_err}) for training either on the SMAL or TOSCA datasets and testing on TOSCA's 286 intra-category animal pairs. For the latter setting, all animal pairs were considered during training without using ground-truth correspondence data.  As seen from the table, our model achieves the best results for both measures in both evaluation settings compared to the other point-based methods.

%\dvir{all  intra-category  animalpairs was chosen for the training on TOSCA, resulting in286 training and test pairs, where no ground-truth data was used during the training of any of the methods}.

We note that the training examples in the SMAL dataset were generated from a parametric model for animals~\cite{zuffi20173dmenagerie}. The model may result in intersecting geometry, such as a leg crossing the body or another leg of the animal. The learning pipeline of spectral correspondence methods, such as SURFMNet~\cite{roufosse2019unsupervised} and GeoFMNet~\cite{donati2020deep}, requires the Cholesky decomposition~\cite{krishnamoorthy2013matrix} of the matrix of features projected on the shape's spectral basis~\cite{litany2017deep}. Unfortunately, in the case of non-watertight or topology-intersected meshes, the Cholesky decomposition is numerically unstable~\cite{higham1990analysis, krishnamoorthy2013matrix, ginzburg2020dual} and the spectral methods cannot be trained on such a dataset. A similar phenomenon occurred for the TOSCA training set. Thus, we did not report results for these methods in table~\ref{tbl:smal_and_tosca_supp}.

%-------------------------------------------------------------------------
\subsection{Additional Visual Comparison} \label{subsec:visual_animal_supp}
Figure~\ref{fig:comp_tosca_supp} presents an additional qualitative comparison for the TOSCA test set. Similar to the findings in the paper, DPC outputs a more accurate matching map between the point clouds compared to the other works.

\input{supplementary/figures/surreal_on_surreal/surreal_on_surreal_pdf.tex}
\input{supplementary/figures/additional_comp_shrec/additional_comp_shrec_pdf.tex}
\input{supplementary/figures/failure_cases/failure_cases_pdf.tex}

\input{supplementary/figures/comparison_tosca/comparison_tosca.tex}

%% file: supplementary/figures/surreal_on_surreal/surreal_on_surreal_pdf.tex
\begin{figure}[tb!]
\begin{center}
\includegraphics[width=\columnwidth]{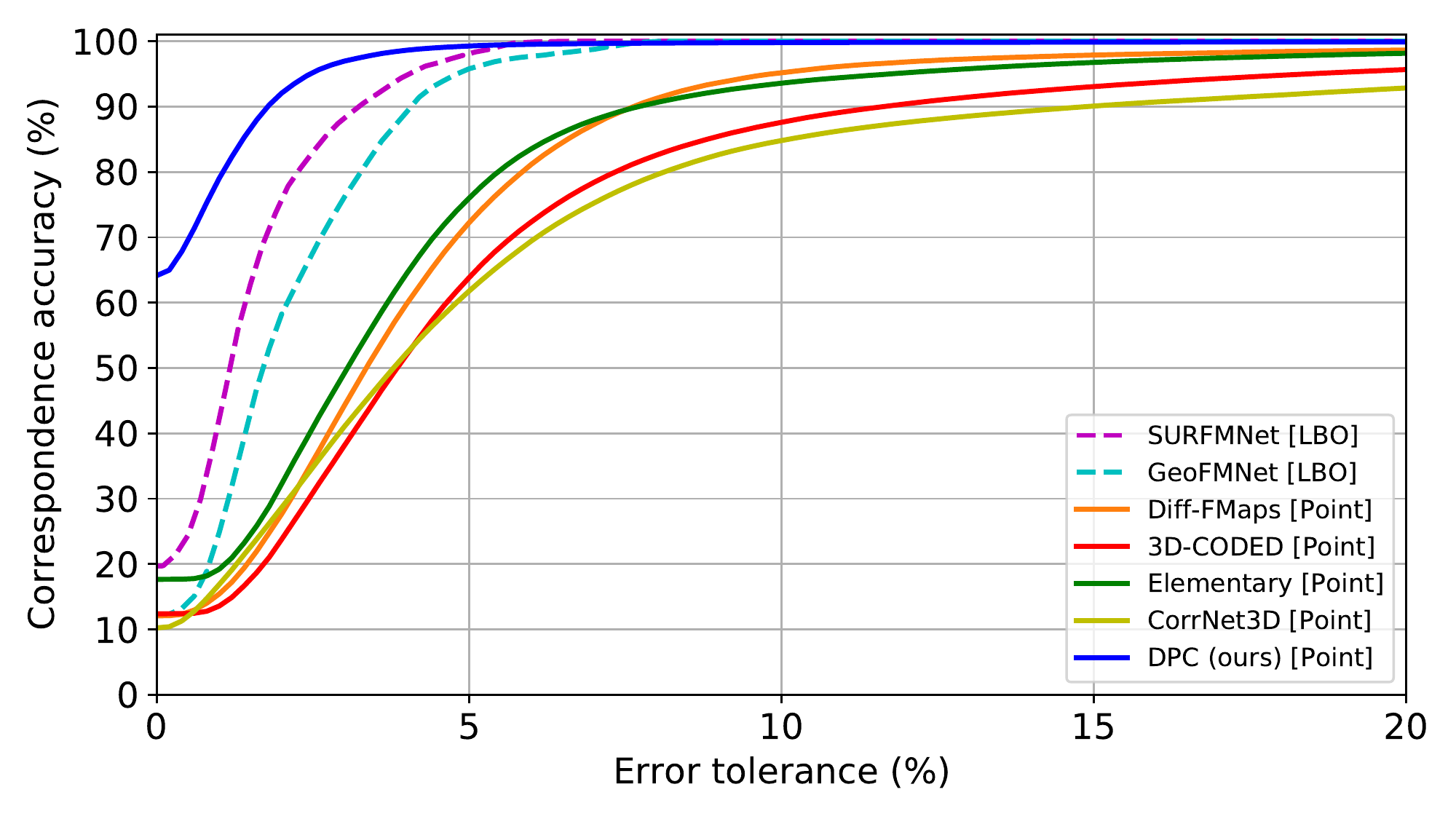}
\caption{{\bfseries Intra-dataset correspondence test.} Training and evaluation was done on SURREAL train and test shape pairs, respectively. The method's type is stated in the squared brackets. Our DPC outperforms the other models, with large performance gains at near-perfect hit rate (\ie, correspondence accuracy for a low error tolerance).}
\label{fig:surreal_on_surreal_supp}
\end{center}
\end{figure}

%% file: supplementary/figures/additional_comp_shrec/additional_comp_shrec_pdf.tex
\begin{figure*}[tb!]
\begin{center}
\begin{tabular}{c c c c c c}
\multicolumn{6}{c}{\includegraphics[width=0.97\linewidth]{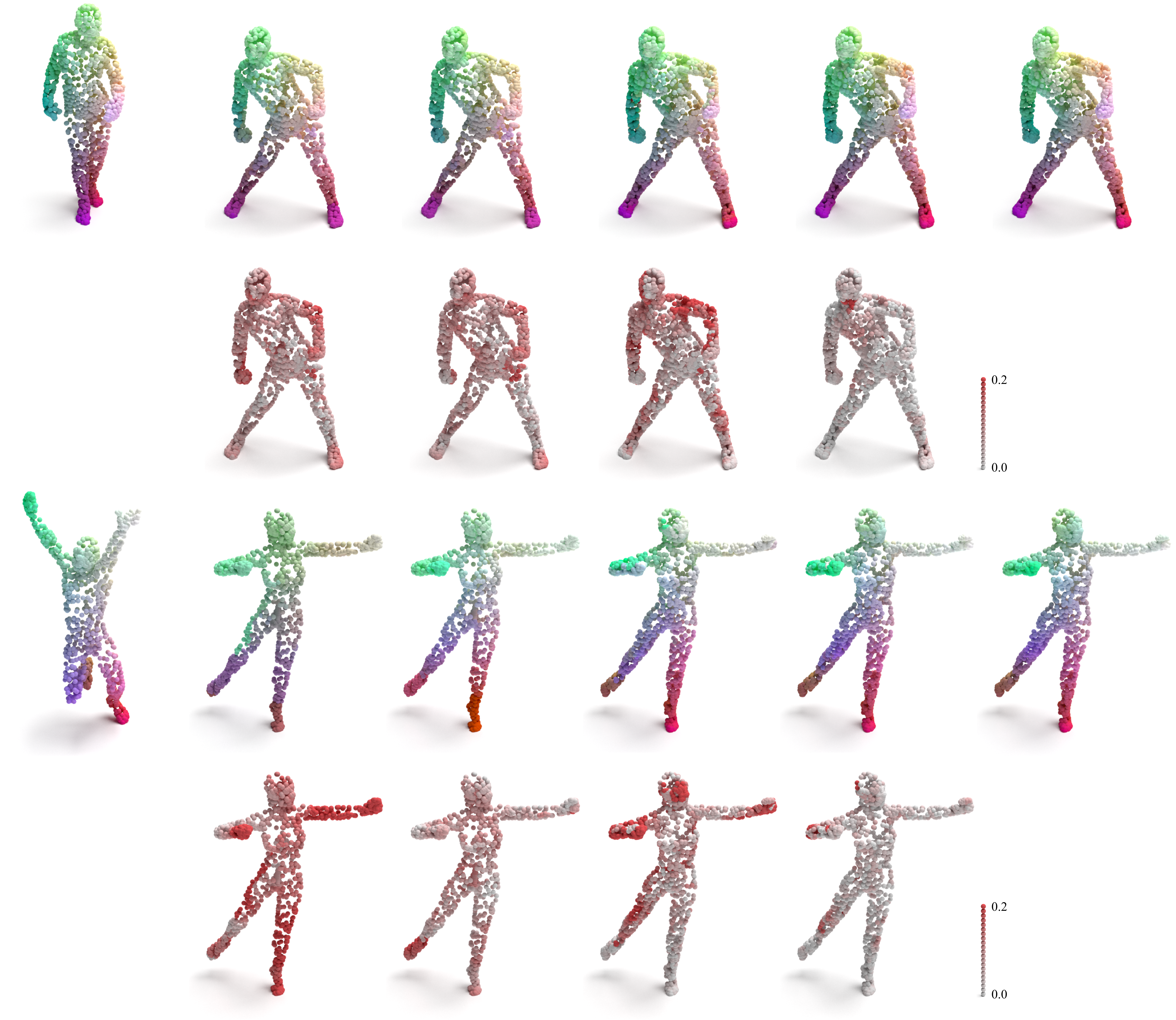}} \\
\whitetext{a}Reference target & \whitetext{}3D-CODED~\cite{groueix20183dcoded} & \whitetext{aa}Elementary~\cite{deprelle2019learning} & \whitetext{aa}CorrNet3D~\cite{zeng2020corrnet3d} & \whitetext{aaa}DPC (ours) & \whitetext{a}Ground-truth \\
\end{tabular}
\end{center}
\caption{\textbf{Additional visual comparison for the SHREC'19 test set.} The training was done on the SURREAL dataset. First and third rows: color-coded point matches. Second and fourth rows: heat-map of correspondence errors magnitude, normalized by the maximal distance between points in the reference target shape. The compared methods result in relatively high errors. In contrast, our DPC better succeeds in aligning the shapes.}
\label{fig:additional_comp_shrec_supp}
\end{figure*}

%The training was done on the SURREAL dataset. Point matches are color-coded. The compared methods result in wrong matches for the palm or foot regions, marked with dashed circles (please zoom-in to notice the difference with respect to the ground truth correspondence map). In contrast, our DPC succeeds in aligning the shapes.

%% file: supplementary/figures/failure_cases/failure_cases_pdf.tex
\begin{figure*}[tb!]
\begin{center}
\begin{tabular}{c c c c c c}
\multicolumn{6}{c}{\includegraphics[width=0.97\linewidth]{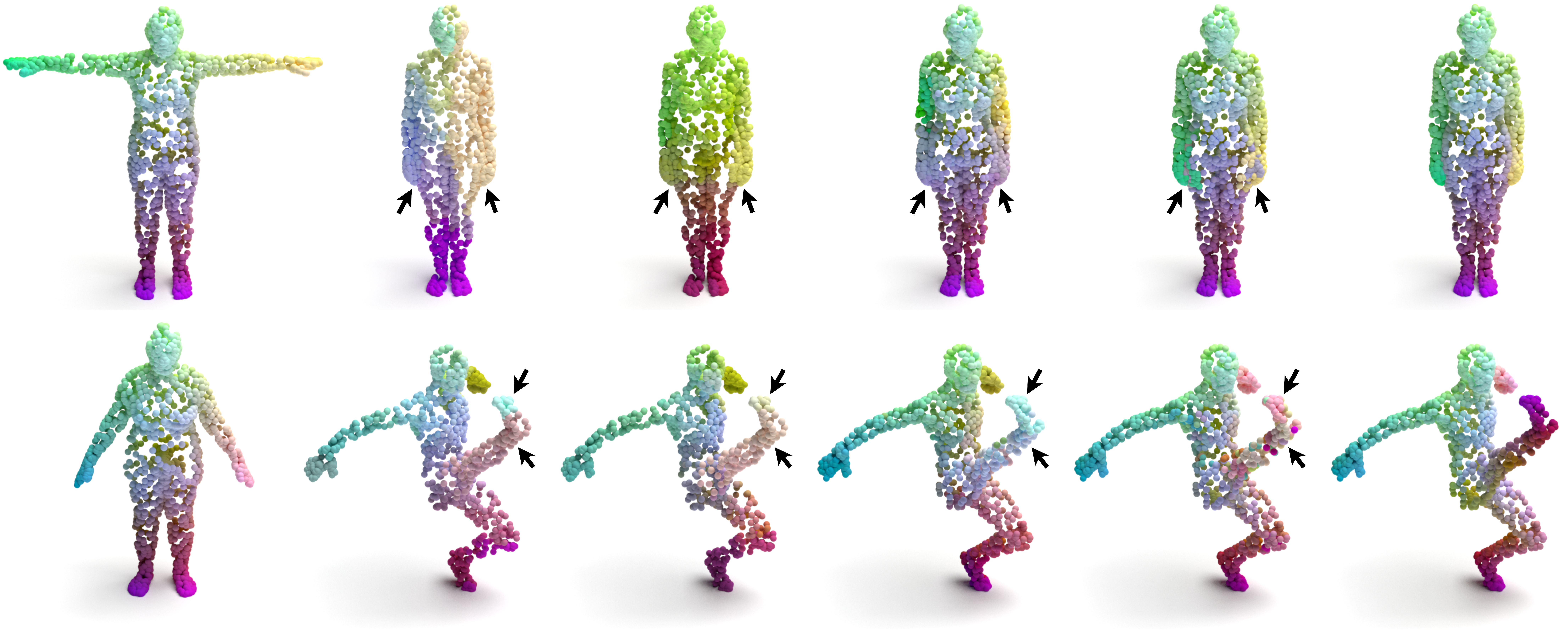}} \\
\whitetext{aaa\:}Reference target & \whitetext{}3D-CODED~\cite{groueix20183dcoded} & \whitetext{a}Elementary~\cite{deprelle2019learning} & \whitetext{a}CorrNet3D~\cite{zeng2020corrnet3d} & \whitetext{aaa}DPC (ours) & \whitetext{aa}Ground-truth \\
\end{tabular}
\end{center}
\caption{\textbf{Failure cases.} We show failure examples of our method for the SHREC'19 test set, where wrong matches are indicated by arrows. For comparison, we include the outcome of other works for these cases as well. The point mappings are color-coded. All the compared methods result in an inaccurate correspondence map for these challenging shape pairs. }
\label{fig:failure_cases_supp}
\end{figure*}

%% file: supplementary/figures/comparison_tosca/comparison_tosca.tex
\begin{figure*}[tb!]
\begin{center}
\begin{tabular}{c c c c c c}
% cat
\includegraphics[width=0.14\linewidth]{figures/comparison_tosca/dpc/s_07_corr_bac.png} & % target gt
\includegraphics[width=0.14\linewidth]{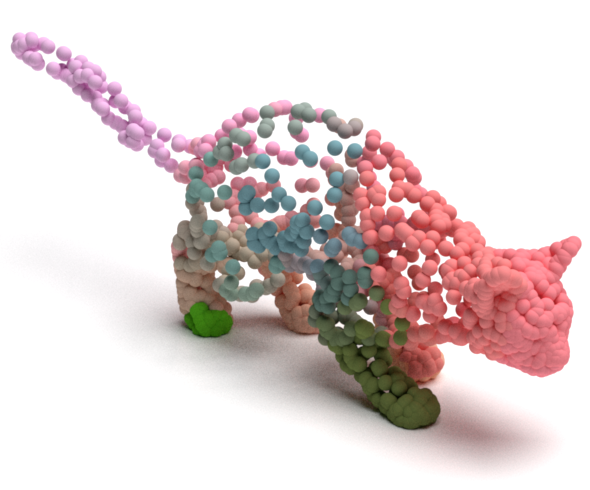} & % 3D-CODED
\includegraphics[width=0.14\linewidth]{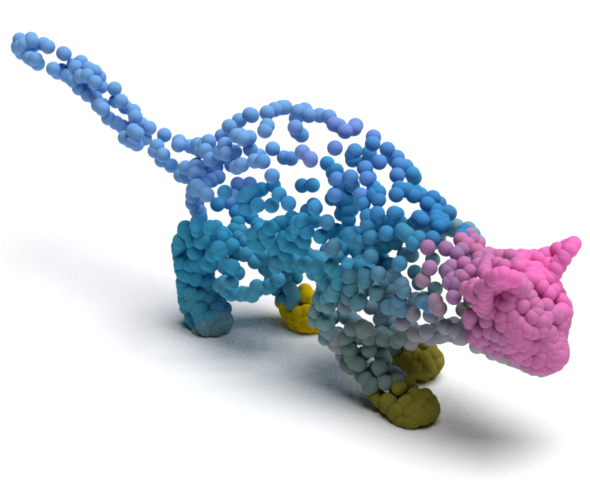} & % Elementary Structures
\includegraphics[width=0.14\linewidth]{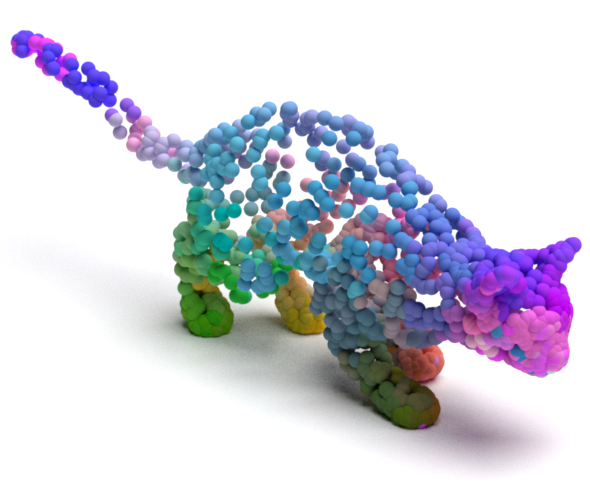} & % CorrNet3D
\includegraphics[width=0.14\linewidth]{figures/comparison_tosca/dpc/t_07_corr_bac.png} & % DPC (ours)
\includegraphics[width=0.14\linewidth]{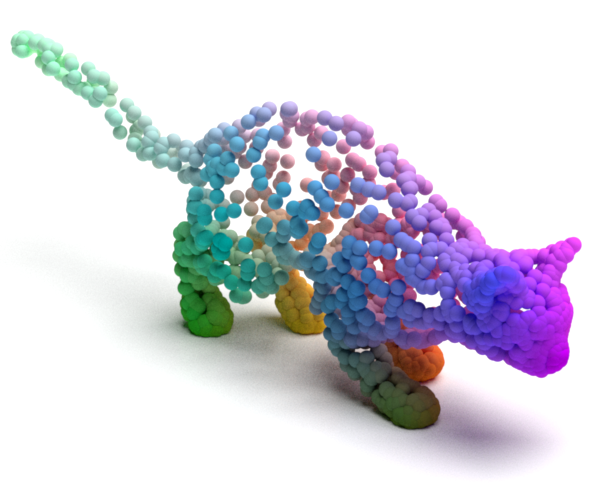} \\ % source gt

% dog
\includegraphics[width=0.14\linewidth]{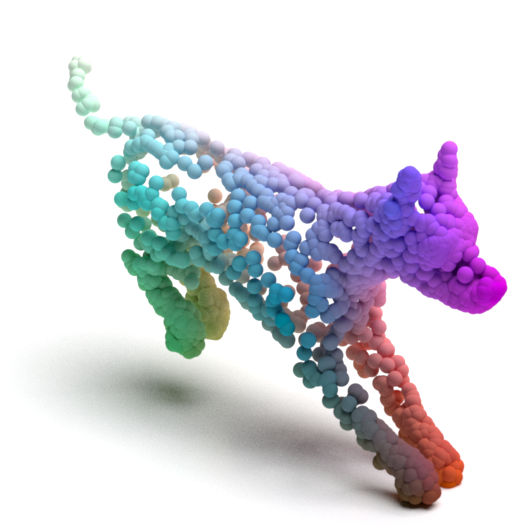} & % target gt
\includegraphics[width=0.14\linewidth]{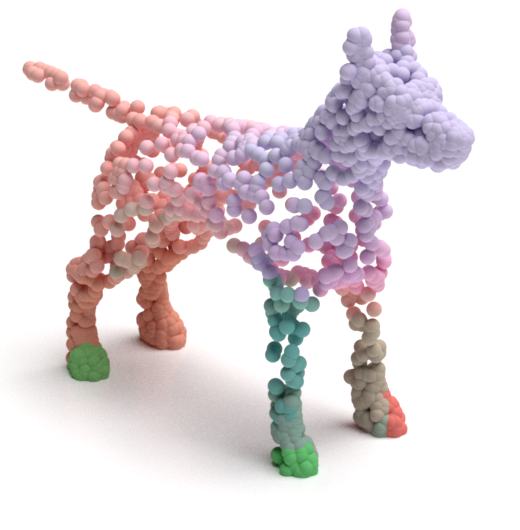} & % 3D-CODED
\includegraphics[width=0.14\linewidth]{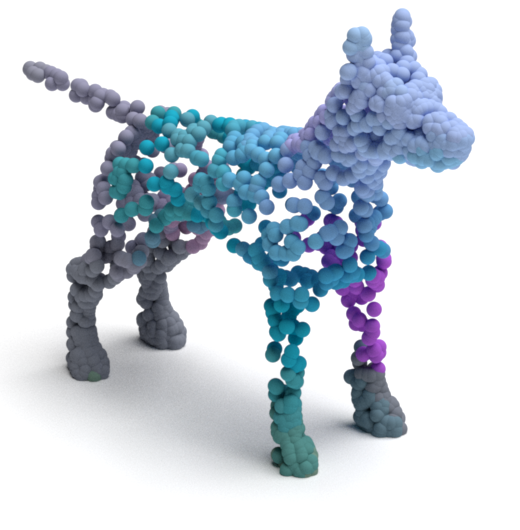} & % Elementary Structures
\includegraphics[width=0.14\linewidth]{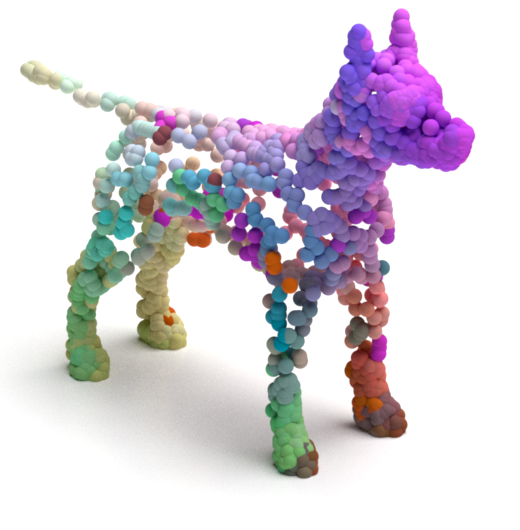} & % CorrNet3D
\includegraphics[width=0.14\linewidth]{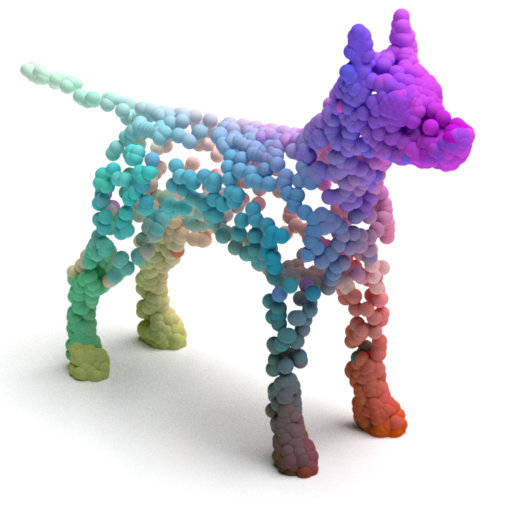} & % DPC (ours)
\includegraphics[width=0.14\linewidth]{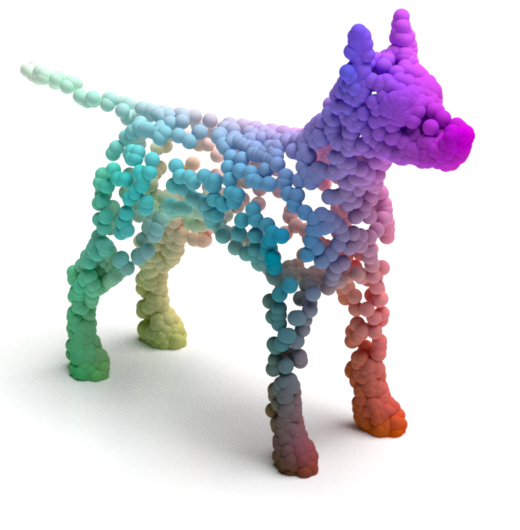} \\ % source gt

Reference target & 3D-CODED~\cite{groueix20183dcoded} & Elementary~\cite{deprelle2019learning} & CorrNet3D~\cite{zeng2020corrnet3d} & DPC (ours) & Ground-truth \\
\end{tabular}
\caption{\textbf{Visual comparison for animal shapes from the TOSCA test set.} The training was done on the SMAL dataset. While the result of the other models is patchy or noisy, our method computes an accurate correspondence result (color-coded).}
\label{fig:comp_tosca_supp}
\end{center}
\end{figure*}

%% file: supplementary/c_ablation_study.tex
\section{Ablation Study} \label{sec:ablation_study_supp}
We verified the design choices in our method by an ablation study, where each time one element in the system was changed and the others were kept the same. The following settings were examined: (a) non-local feature aggregation in the point embedding module (as in the original DGCNN~\cite{wang2019dynamic} model) instead of aggregation from local point neighbors; (b) unbounded dot-product similarity (the numerator of Equation~\ref{eq:similarity}) instead of the cosine similarity; (c) considering all neighbors for the cross-construction operation ($k_{cc}=n$); (d) excluding the self-construction module ($\lambda_{sc} = 0$); and (e) turning off the mapping loss ($\lambda_m = 0$). Table~\ref{tbl:ablation_study_supp} summarizes the results.

The table indicates the necessity of the proposed components and their configuration, as all the ablative settings compromise the model's performance. Local feature aggregation enables the model to extract a discriminative point representation and the bounded cosine similarity contributes to the numerical stability of the learning process. Additionally, the ablative experiments validate that a local latent neighborhood for the cross-construction operation and the employment of the self-construction module are highly important considerations in our method.

%In addition, while considering all target points for a source point mapping is a common approach in correspondence works~\cite{wang2019deep, zeng2020corrnet3d}, it is ineffective in our case. Instead, for each source point we consider a local latent neighborhood in the target shape. It enables the model to explore relevant candidates for matching. The ablation study also shows that it is better to find neighbors in the feature space for self-construction rather than in the Euclidean space. Adjacent Euclidean points for non-rigid shapes may be far apart in the underling surface, which motivate the latent neighbor selection instead. Finally, the ablation experiments indicates the importance of both self and cross-construction losses for the training of our DPC model.

%% file: supplementary/d_experimental_settings.tex
\section{Experimental Settings} \label{sec:experimental_settings_supp}

\subsection{Feature Extraction Architecture} \label{subsec:architecture_supp}

We adopt the DGCNN architecture~\cite{wang2019dynamic}, with a local point neighborhood rather than a dynamic non-local one. The network includes 4 per-point convolution layers with filter sizes $(96, 192, 384, 768)$. Batch normalization and a leaky ReLU activation with a negative slope of $-0.2$ are used after each filer. The convolutions operate on the concatenation of the point features and its edge features, where the latter are differences between the features of the points and its 27 nearest Euclidean points. After each layer, the per-point features are max-pooled from the point's neighbors. Finally, the features from the different stages are concatenated to a vector of size $1440 = 96 + 192 + 384 + 768$, which is passed through two last layers with $(1044, 512)$ neurons, along with batch normalization and non-linearity as before, to produce a $c = 512$ dimensional feature vector for each point.

\subsection{DPC Optimization} \label{sec:optimization_supp}
Table~\ref{tbl:optimization_supp} summarizes the optimization parameters for our model. The same values were used for all the four training datasets (SURREAL, SHREC'19, SMAL, and TOSCA). We used an Adam optimizer with an initial learning rate of 0.0003, momentum 0.9, and weight decay of 0.0005. The learning rate is multiplied by a factor of 0.1 at epochs 6 and 9. The training was done on an NVIDIA RTX 2080Ti GPU. %and took \redtext{$A$}, \redtext{$B$}, \redtext{$C$}, and \redtext{$D$} hours for the SURREAL, SHREC'19, SMAL, and TOSCA datasets, respectively.

%Table~\ref{tbl:optimization_supp} summarizes the optimization parameters of our correspondence method. We used an Adam optimizer with momentum \redtext{$X$} and different number of epochs, depending on the training dataset. The training was done on a NVIDIA RTX 2080Ti GPU and took between \redtext{$A$} to \redtext{$B$} hours, dependin of the training dataset.

\input{supplementary/tables/smal_and_tosca/acc_and_error.tex}

\input{supplementary/tables/ablation_study/ablation_study.tex}

\input{supplementary/tables/optimization_parameters/optimization_parameters}

%% file: supplementary/tables/smal_and_tosca/acc_and_error.tex
\begin{table}[tb!]
%\small
\centering
\begin{tabular}{ @{ } l c c c c @{} }
\toprule
       & \multicolumn{2}{c}{\textbf{SMAL/}} & \multicolumn{2}{c}{\textbf{TOSCA/}} \\
       & \multicolumn{2}{c}{\textbf{TOSCA}}    & \multicolumn{2}{c}{\textbf{TOSCA}}  \\
       \cmidrule(lr){2-3} \cmidrule(lr){4-5}
Method & \textit{acc} $\uparrow$ & \textit{err} $\downarrow$ & \textit{acc} $\uparrow$ & \textit{err} $\downarrow$\whitetext{a}\: \\
\midrule
3D-CODED~\cite{groueix20183dcoded}     & 0.5\% & 19.2 & *      & *           \\
Elementary~\cite{deprelle2019learning} & 0.5\% & 13.7 & *      & *           \\
CorrNet3D~\cite{zeng2020corrnet3d}     & 5.3\% & 9.8 & 0.3\%  & 32.7         \\
DPC (ours) & \textbf{33.2\%} & \textbf{5.8} & \textbf{34.7\%} & \textbf{2.8} \\
\bottomrule
\end{tabular}
\vspace{0.1cm}
\caption{\textbf{Accuracy and error.} We evaluate the accuracy at 1\% tolerance (\textit{acc}, in percentage) and the average correspondence error (\textit{err}, in centimeters) for two train/test settings of animal datasets. Higher accuracy and lower error reflect a better result. The training on TOSCA was done without correspondence supervision, thus, the supervised techniques~\cite{groueix20183dcoded, deprelle2019learning} were not applied in this setting. Our model achieves better results compared to the competing methods.}
\label{tbl:smal_and_tosca_supp}
\end{table}

% normalization factor of the average error (err):
% head to bottom length (without a tail) of the first cat in the dataset:
% l_data = 0.49+0.89 = 1.28
%
% average cat length:
% l_avg = 0.46
%
% normalization factor:
% norm = 1.28/0.46 = 3

% results before normalization (only the average error (err) is normalized):
%3D-CODED~\cite{groueix20183dcoded}     & 0.5\% & 57.7 & *      & *             \\
%Elementary~\cite{deprelle2019learning} & 0.5\% & 41.2 & *      & *             \\
%CorrNet3D~\cite{zeng2020corrnet3d}     & 5.3\% & 29.3 & 0.3\%  & 98.2          \\
%DPC (ours) & \textbf{33.2\%} & \textbf{17.3} & \textbf{34.7\%} & \textbf{8.5} \\

%% file: supplementary/tables/ablation_study/ablation_study.tex
\begin{table}[tb!]
\centering
\begin{tabular}{ @{ } l c c @{ } }
\toprule
Setting & \textit{acc} $\uparrow$ & \textit{err} $\downarrow$ \\
\midrule
(a) Non-local feature aggregation              & 13.5\% & 6.4 \\
(b) Dot-product similarity measure             & 12.0\% & 6.2 \\
(c) All cross neighbors ($k_{cc} = n$)         & 2.2\% & 7.4  \\
% All self neighbors                           & 1.5\% & 8.8  \\
% Euclidean self neighbors                     & 12.3\% & 6.9 \\
(d) W/O self construction ($\lambda_{sc} = 0$) & 3.4\% & 6.6  \\
(e) W/O mapping loss ($\lambda_m = 0$)         & 11.4\% & 6.7  \\
% W/o cross construction ($\lambda_{cc} = 0$)  & 13.8\% & 6.6 \\
Our complete method                                & \textbf{17.7\%} & \textbf{6.1} \\
\bottomrule
\end{tabular}
\vspace{0.1cm}
\caption{\textbf{Performance in ablative settings.} We train our method on the SURREAL dataset and test it on the SHREC'19 benchmark. The evaluation metrics are the same as in Table~\ref{tbl:surreal_and_shrec} in the paper. The complete proposed method yields the best performance. Please see additional details in Section~\ref{sec:ablation_study_supp}.
}
\label{tbl:ablation_study_supp}
\end{table}

%We verify the design choices in our method by evaluation its accuracy at 1\% tolerance and its average error in different settings, where each time one component is changed. The full method yields the best performance.

%% file: supplementary/tables/optimization_parameters/optimization_parameters.tex
\begin{table}[tb!]
\centering
\begin{tabular}{ @{ } l l c @{ } }
\toprule
&  Description & Value \\
\midrule
$k_{cc}$ & Cross-construction neighborhood size & 10 \\
$k_{sc}$ & Self-construction neighborhood size & 10 \\
$k_m$ & Mapping loss neighborhood size &10 \\
$\alpha$ & Mapping loss neighbor sensitivity & 8 \\
$\lambda_{cc}$ & Cross-construction loss weight & 1 \\
$\lambda_{sc}$ & Self-construction loss weight & 10 \\
$\lambda_m$ & Mapping loss weight & 1 \\
BS & Batch size & 8 \\
LR & Learning rate & 0.0003 \\
TEs & Training epochs & 300 \\
\bottomrule
\end{tabular}
\vspace{0.1cm}
\caption{\textbf{Hyper-parameters.} The tables details the hyper-parameter values that we used for the training of DPC.}
\label{tbl:optimization_supp}
\end{table}

%The tables details the hyper-parameter values that we used for the training of DPC on different datasets. BS, LR, TEs and TT stand for batch size, learning rate, training epochs, and training time, respectively. The training time is reported in hours for an NVIDIA RTX 2080Ti GPU.